\documentclass[journal]{IEEEtran}

\usepackage[version=0.96]{pgf}
\usepackage{tikz}
\usetikzlibrary{arrows,shapes,snakes,automata,backgrounds,petri}
\usepackage[latin1]{inputenc}
\usepackage{verbatim}

\usepackage{tabularx}
\usepackage{enumitem}
\usepackage{multirow}
\usepackage{booktabs}
\usepackage{caption}
\usepackage{threeparttable}

\usepackage{bm}
\usepackage{amsfonts}
\usepackage{amsmath}
\usepackage[colorlinks]{hyperref}
\usepackage{breqn}

\usepackage{url}
\usepackage{stackengine}
\usepackage{float}
\usepackage{array}
\usepackage{algorithm}
\usepackage{algorithmic}

\usepackage{soul}
\usepackage{color}
\usepackage{cleveref}

\usepackage{rotating}

\definecolor{Orange}{rgb}{0.9,0.5,0}
\definecolor{NavyBlue}{rgb}{0.1, 0.4, 0.8}
\definecolor{Magenta}{rgb}{0.8, 0.1, 0.6}
\definecolor{mypink2}{RGB}{219, 48, 122}

\newcommand{\change}[1]{#1}

\ifCLASSINFOpdf
\else
\fi

\begin{document}

\title{ISF-GAN: An Implicit Style Function for High-Resolution Image-to-Image Translation}

\author{Yahui~Liu,
        Yajing~Chen,
        Linchao~Bao,
        Nicu~Sebe~\IEEEmembership{Senior Member,~IEEE}, 
        Bruno~Lepri,
        and~Marco~De~Nadai %
\thanks{Y. Liu and N. Sebe are with the Department of Information Engineering and Computer Science, University of Trento, Italy. E-mail: \{yahui.liu, niculae.sebe\}@unitn.it.}%
\thanks{Y. Chen and L. Bao are with Tencent AI Lab, Shenzhen, China. E-mail: \{jadeyjchen, linchaobao\}@tencent.com.}
\thanks{B. Lepri and M. D. Nadai are with Fondazione Bruno Kessler, Italy. E-mail: lepri@fbk.eu. and work@marcodena.it}
\thanks{This work is done during internship at Tencent AI Lab, Shenzhen, China.}
}

\markboth{Manuscript under review.}%
{Shell \MakeLowercase{\textit{et al.}}: Bare Demo of IEEEtran.cls for IEEE Journals}

\maketitle

\begin{abstract}
Recently, there has been an increasing interest in image editing methods that employ pre-trained unconditional image generators (e.g., StyleGAN). However, applying these methods to translate images to multiple visual domains remains challenging. Existing works do not often preserve the domain-invariant part of the image (e.g., the identity in human face translations), or they do not usually handle multiple domains or allow for multi-modal translations.
This work proposes an implicit style function (ISF) to straightforwardly achieve multi-modal and multi-domain image-to-image translation from pre-trained unconditional generators.
The ISF manipulates the semantics of a latent code to ensure that the image generated from the manipulated code lies in the desired visual domain. Our human faces and animal image manipulations show significantly improved results over the baselines.
Our model enables cost-effective multi-modal unsupervised image-to-image translations at high resolution using pre-trained unconditional GANs. The code and data are available at: \url{https://github.com/yhlleo/stylegan-mmuit}.
\end{abstract}


\IEEEpeerreviewmaketitle

\section{Introduction}
\label{sec:introduction}

\begin{figure}[ht]	
	\renewcommand{\tabcolsep}{1pt}
	\centering
	\begin{tabular}{c}
		\includegraphics[width=\linewidth]{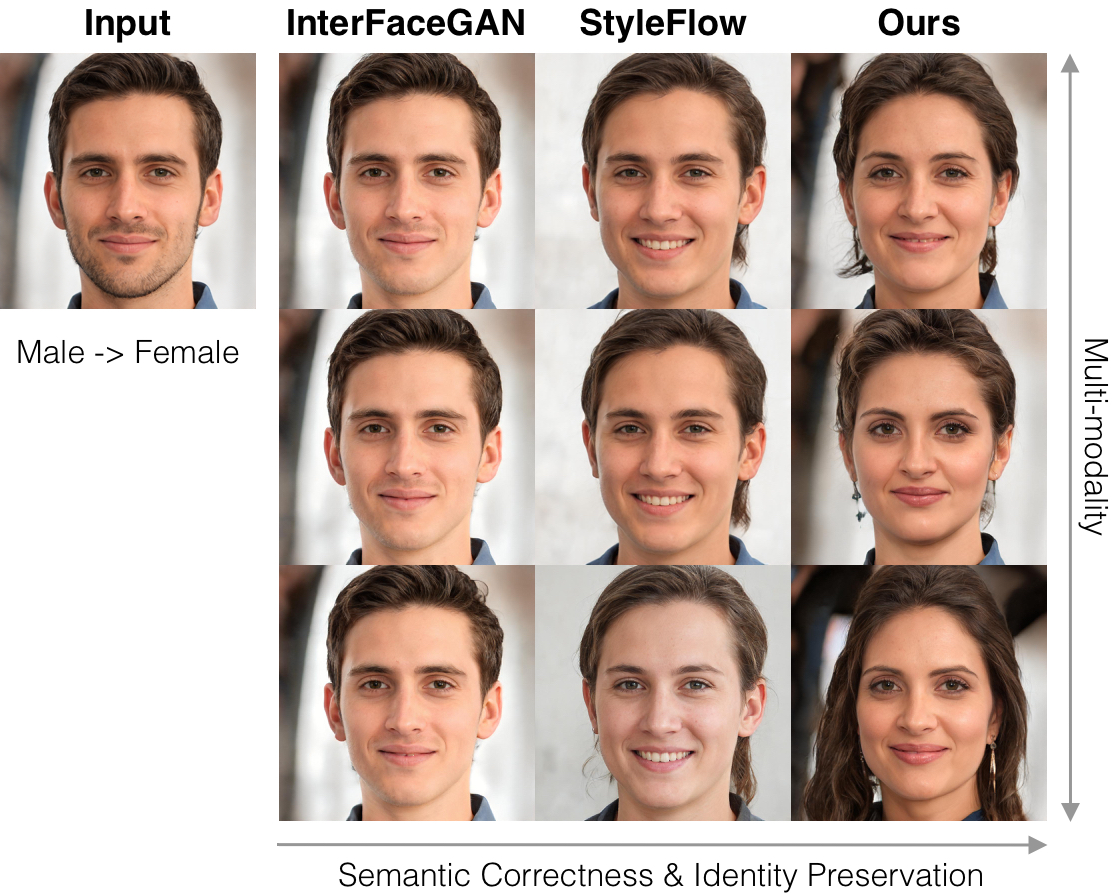}
	\end{tabular}
	\caption{
	\change{Our model focuses on Multi-modal and Multi-domain Unsupervised Image-to-image Translation. In this figure, we show an male$\rightarrow$female translation, in which we wish to change the gender of the input image without changing some facial features that allows us to recognize that the input image and the output image depict the same person. Additionally, we want to generate multiple diverse images for each translation.
	We can observe that state-of-the-art models based on StyleGAN, do not usually maintain the content of the original image (e.g., background and people's identity), do not generate images with correct semantics, and have limited diversity. Our model better adheres to these properties.}
	} 
	\label{Fig:teaser}
\end{figure}

Generative methods have become increasingly effective at synthesizing realistic images at high resolution, stimulating new practical applications in academia and industry. 
Many different tasks have been proposed, from super-resolution~\cite{ledig2017photo, menon2020pulse} to image manipulation~\cite{lee2018context, jo2019sc}, from image-to-image and text-to-image translations~\cite{choi2018stargan, choi2019stargan, li2019controllable, liu2020gmm} to video generation~\cite{siarohin2020first,tulyakov2018mocogan}. 
Yet, training these task-specific models at high resolution (e.g., $1024 \times 1024$) is very computationally expensive.
\change{For this reason, recent works have interpreted and exploited the latent space of pre-trained high-resolution unconditional Generative Adversarial Networks (GANs) to solve several generative tasks without training a generator from scratch~\cite{abdal2019image2stylegan, abdal2020image2stylegan2,abdal2020styleflow, abdal2021labels4free, pan20202d, tian2021good, zhang2020image}. 
Most of these approaches are based on StyleGAN~(\cite{karras2019style,karras2020analyzing}), the state-of-the-art of unconditional image generation. 
StyleGAN maps a noise vector $\bm{z} \sim \mathcal{N}(\bm{0}, \mathbf{I}_n)$ to an intermediate and learned latent space $\bm{\mathcal{W}}^+$, which exhibits some intriguing disentangled semantic properties~\cite{collins2020editing, harkonen2020ganspace, wu2021stylespace} that can be interpreted and exploited.
For example, InterFaceGAN~\cite{shen2020interpreting,shen2020interfacegan} identifies the semantic attributes in $\bm{\mathcal{W}}^+$ to manipulate the semantics of a latent code and change the facial attributes of images.
However, existing models manipulating StyleGAN latent codes allow to edit just one attribute per time or fail to preserve the content of the image not involved in the manipulation, resulting, for example, in eye color changes when gender is manipulated (see \Cref{Fig:teaser}). Moreover, most existing models are deterministic (i.e., not multi-modal). Altogether, these issues make existing approaches unsuitable for Multi-modal, and Multi-domain Unsupervised Image-to-image Translation (MMUIT)~\cite{choi2019stargan, Liu_2021_CVPR}.}

\change{An ideal MMUIT model should be able to change the \emph{domain}-specific parts of the image while preserving the \emph{content} of the image (e.g., the background and the identity of people) and synthesize images considering the multiple appearance \emph{modes} within each domain~\cite{choi2019stargan, Liu_2021_CVPR}.
Here, \emph{domain} refers to a set of images having some distinctive visual pattern, usually called \emph{style}. For example, we can group images based on the gender of people in pictures (see \Cref{Fig:teaser}).
Since paired images (e.g., pictures of the same person with different gender) are usually
not available, the model should be unsupervisedly trained~\cite{choi2018stargan, choi2019stargan, Liu_2021_CVPR, huang2018multimodal}.}

This paper proposes an implicit style function (ISF) that allows leveraging a pre-trained unconditional image generator to do MMUITs.
The proposed method works directly on the latent space and enables the change of multiple semantic attributes at a time without affecting any content of the image that is not involved in the translation. 
Given a latent code in the latent space of pretrained GANs (e.g., $\bm{\mathcal{W}}^+$ in StyleGAN~\cite{karras2019style}), also called style code, $\bm{w}$, a semantic attribute label $\bm{d}$, and a noise vector $\bm{z} \sim \mathcal{N}(\bm{0}, \mathbf{I}_n)$, the ISF outputs a new style code $\bm{w}*$ that contains the desired semantic attributes.
By randomly sampling multiple $\bm{z}$, several variants of the desired image translation (e.g., multiple hairstyles in gender manipulations) can be generated.

As a consequence, our proposed method can be used for multiple tasks, including Multi-domain Multi-modal Unsupervised Image-to-image Translation~\cite{choi2019stargan, lee2019drit++} and smooth semantic interpolation~\cite{chen2019homomorphic}. When used in conjunction with GAN \emph{inversion} techniques, which find the most similar style code given an input image (e.g.,~\cite{abdal2020image2stylegan2, richardson2020encoding}), our proposed method can also translate real images at high-resolution. 

To enable better manipulations of StyleGAN's latent codes, we propose a simple extension of Adaptive Instance Normalization (AdaIN)~\cite{huang2017arbitrary}, namely Adaptive Layer Normalization (AdaLN).
AdaLN normalizes
the latent code through Layer Normalization~\cite{ba2016layer} instead of Instance Normalization~\cite{ulyanov2016instance} and goes beyond the assumption of independence between channels by computing the normalization for each sample across all the channels. 
Quantitative and qualitative results in face manipulations show that our proposed method outperforms state-of-the-art methods.

Our contributions can be summarized as follows: 
\begin{itemize}
    \item We propose an implicit style function for editing multiple semantics of the input image at once by StyleGAN latent space manipulations. The images generated from the edited latent codes preserve the content/identity of input images while changing the user-specified semantics;
    \item We propose AdaLN, a simple extension of the popular normalization method AdaIN~\cite{huang2017arbitrary} to enable better arbitrary style transfer in the StyleGAN latent space; 
    \item At the best of our knowledge, we are the first to address multi-domain and multi-modal unsupervised image-to-image translation by a \textit{pre-trained} and \textit{fixed} image generator, which allows having image translations at high resolution (i.e., 1024$\times$1024).
\end{itemize}

\section{Related Work}
Unsupervised MMUIT models aim at learning a mapping between different visual domains unsupervisedly.
They require that: 1) the domain-invariant part of the image, also called \emph{content}, is preserved; 2) the model can map multiple domains; and 3) the output is multi-modal (i.e., diverse translations can be generated with the same input image).

To the best of our knowledge, there are no works, based on pre-trained and fixed unconditional GANs, that simultaneously meet all the requirements of unsupervised MMUIT models. However, we here review the literature of MMUIT models, StyleGAN-based latent space manipulations and StyleGAN-based inversion techniques.

\label{sec:related-work}
\noindent\textbf{Image-to-image Translation.}
Early attempts in Image-to-image Translation are based on paired images~\cite{isola2017image,zhu2017toward,aliaksandr2018Deformable,hoffman2018cycada} and one-to-one domain mappings~\cite{zhu2017unpaired,huang2018multimodal, lee2018diverse, pumarola2018ganimation, mao2019mode, wang2020deepi2i}. 
More recently, many studies have focused on training a single model that maps images into multiple domains and overcomes the deterministic (i.e., one-to-one) translation often assumed by previous works (e.g.,~\cite{choi2018stargan, liu2017unsupervised, pumarola2018ganimation}).
We refer to these methods as MMUIT models.
For example, 
DRIT++~\cite{lee2019drit++} assumes the existence of a domain-independent (``content'') and a domain-specific (``style'') image representation and obtain this using two separate encoders. Then, DRIT++ allows multi-modal translations injecting random noise in the generation process. 
GMM-UNIT~\cite{liu2020gmm,liu2020describe} uses a Variational Auto-Encoder approach~\cite{kingma2013auto} where a style encoder maps the image into a Gaussian Mixture Model, from which it is possible to sample multiple different styles to be used in the generation process.
StarGAN v2~\cite{choi2019stargan,Liu_2021_CVPR} proposes an architecture composed by a style encoder, which maps an existing image to a style code, and a mapping function, which starts from random noise and a domain code to generate a style code. In doing so, they achieve state-of-the-art performance in high-resolution (i.e., $256 \times 256$) and diversity of translated images.

The current state-of-the-art models in MMUIT field show results usually at  $256\times 256$ resolution and require training a generator from scratch. We here aim at enabling high-resolution MMUIT without training a generator.

\noindent\textbf{StyleGAN Latent Space Manipulation.} 
Most conditional GANs are trained at low and medium resolutions (i.e., up to $256\times 256$ pixels). Scaling the image generators to higher resolutions to solve specific tasks requires a substantial computational budget.
For this reason, scholars have recently started exploring and interpreting the latent code semantics of large unconditional networks for image synthesis such as PGGAN~\cite{karras2018progressive} and StyleGAN~\cite{karras2019style, karras2020analyzing}. 
The general idea is to adapt unconditional GANs to solve specific tasks at high resolution without training task-specific generators~\cite{tian2021good}.
For example, H{\"a}rk{\"o}nen \emph{et al.}~\cite{harkonen2020ganspace} identify important latent directions based on Principal Component Analysis (PCA) to control properties such as lighting, facial attributes, and landscape attributes.
Shen \emph{et al.}~\cite{shen2020interfacegan} use an off-the-shelf classifier to find the linear hyperplanes of semantic facial attributes.
InterFaceGAN~\cite{shen2020interpreting,shen2020interfacegan} learns the latent semantics through linear hyper-planes, which rely on multiple SVMs that have to be specifically trained on each domain translation. Then, it edits face images moving the latent code with linear transformations.
PSP~\cite{richardson2020encoding} learns an encoder and a mapping function that enables multi-modal domain translations but does not support multi-domain translation. 
Chai \emph{et al.}~\cite{chai2021using} propose to learn an \emph{imperfect} masked encoder that finds the latent code that best resembles the input masked or coarsely-edited image. Then, it uses the fixed generator to synthesize photo-realistic results showing applications in image composition, image inpainting and multi-modal editing. Finally, StyleFlow~\cite{abdal2020styleflow} proposes to have attribute-controlled sampling and attributed-controlled editing through StyleGAN. They learn a function $\Phi(z,d)$ to sample StyleGAN latent codes conditioned to a semantic attribute vector, and they allow editing specific blocks of the StyleGAN latent space to edit pre-defined semantic categories.

However, existing latent space manipulation models based on StyleGAN are not suitable for MMUIT tasks. They are either limited on single-attribute at a time  (e.g.,~\cite{shen2020interfacegan, shen2020interpreting}) or one-domain translations (e.g.,~\cite{richardson2020encoding}). Moreover, they have very limited if no diversity in the manipulations (e.g.,~\cite{shen2020interfacegan, shen2020interpreting, richardson2020encoding}) and, more importantly, manipulations along one attribute can easily result in unwanted changes along with other attributes (e.g., face features changes in hair color manipulations).
We propose a simple and effective Implicit Style Function to manipulate the StyleGAN latent code and enable MMUIT at high resolution with a pre-trained and fixed StyleGAN model.

\noindent\textbf{GAN inversion in StyleGAN latent space.} 
Solving the GAN inversion problem is essential to use pre-trained GANs with real images.
GAN inversion techniques aim to find the generator's latent code that best corresponds to a given image. 

Existing inversion approaches typically fall into two categories: encoder-based and optimization-based methods. 
The former~\cite{richardson2020encoding} uses LBFGS~\cite{liu1989limited} or similar optimizers to find the latent code $\bm{z}$ that best recovers the image $\bm{x}$ with $\bm{z}^* = \arg\min_{\bm{z}} (\text{dist}(G(\bm{z}), \bm{x}))$ where $\text{dist}$ is a metric distance function in the image space.
The latter models (e.g.,~\cite{abdal2019image2stylegan,abdal2020image2stylegan2,guan2020collaborative}) instead speeds-up the process at inference time by learning an encoder $E$. Then, the optimal $\bm{z}$ is simply the result of a feed-forward pass through $E$, more formally: $\bm{z}^* = E(x)$. 

\noindent\textbf{Fusion methods.} 
A noteworthy application of Image-to-image Translation is image fusion, which aims to exploit images obtained by different sensors to generate better and more robust images.
For example, FusionGAN~\cite{ma2019fusiongan} focuses on fusing the thermal radiation information in infrared images and the texture detail information in visible images to generate images through a GAN.
Pan-GAN~\cite{ma2020pan} instead focuses on fusing the thermal radiation information in infrared images and the texture detail information in visible images to generate images through a GAN. The latter focuses on remote sensing images fusing different layers of information to do pan-sharpening and better preserve the spectral and spatial information in images.

\section{Method}
\label{sec:method}

\begin{figure*}[ht]	
	\renewcommand{\tabcolsep}{1pt}
	\centering
	\begin{tabular}{c}
	    \includegraphics[width=.82\linewidth]{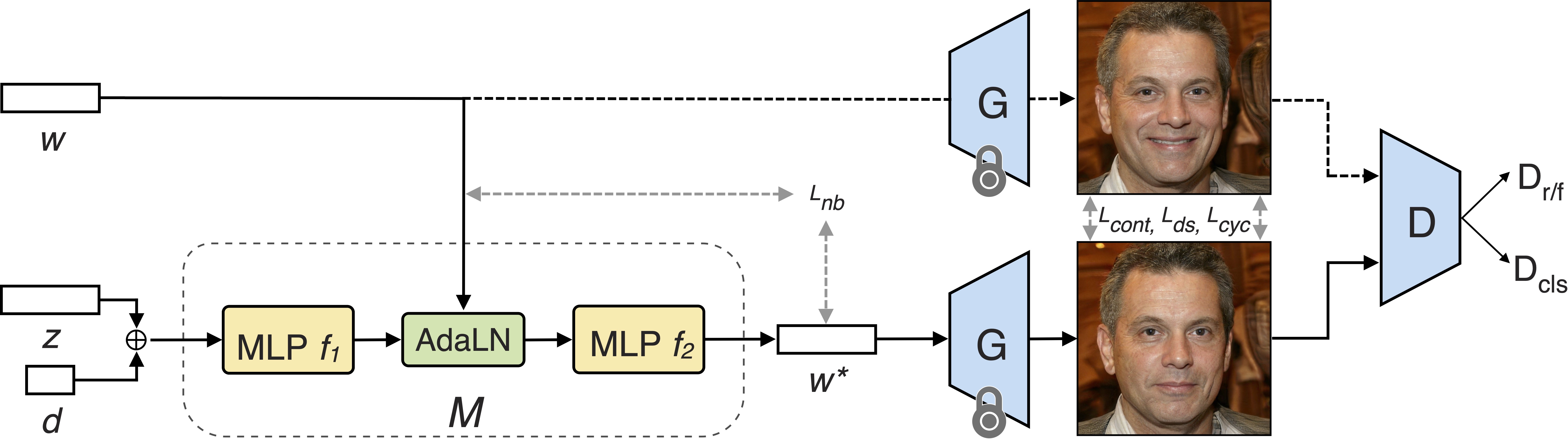}
	\end{tabular}
	\caption{We train an Implicit Style Function $\mathcal{M}$ that manipulates a style code $\bm{w}$ into $\bm{w}^*$ given a randomly sampled noise $\bm{z}$ and a semantic attribute label $\bm{d}$.
	$\mathcal{M}$ is trained so that the image generated by $G$ should have the semantics specified by $\bm{d}$ without changing anything along other attributes (e.g., face identity).
	$G$ is a \emph{pre-trained} and \emph{fixed} uncodintional GANs (e.g., StyleGAN). We also train a discriminator $D$ that discriminates between real/fake images and classifies the image attributes. 
	} 
	\label{Fig:framework}
\end{figure*}

\label{sec:semantics}
An unconditional generator $G: \bm{\mathcal{Z}} \rightarrow \bm{\mathcal{X}}$ learns to synthesize an image $\bm{x} \in \bm{\mathcal{X}}$ given a vector $\bm{z} \in \bm{\mathcal{Z}}$, lying in a low-dimensional latent space.
In this work, we focus on StyleGAN v1~\cite{karras2019style} and StyleGAN  v2~\cite{karras2020analyzing}, which first learn a non-linear mapping $f: \bm{\mathcal{Z}} \rightarrow \bm{\mathcal{W}}^+$ that outputs $\bm{w} = f(\bm{z}) \in \mathbb{R}^{18\times512}$, and then generates the image from $\bm{\mathcal{W}}^+$ space: $\bm{x} = G(\bm{w})$.

We aim at manipulating $\bm{w}$ such that it follows some user-specified semantic attributes $\bm{d} \in \bm{\mathcal{D}}\subset\mathbb{R}^m$ that refer to a $m$-dimensional semantic space, while best preserving the attribute-invariant content (e.g., face identity) of the source images.
\change{In other words, we want to learn a non-linear neural Implicit Style Function (ISF) $\mathcal{M}: \bm{\mathcal{W}}^+ \rightarrow \bm{\mathcal{W}}^+$ that outputs a manipulated code $\bm{w}^*$ in which:
\begin{enumerate}
    \item \textit{The output looks realistic.} The generated image $\bm{x}^* = G(\bm{w}^*)$ should look as realistic as the images generated from non-manipulated latent codes;
    \item \textit{Generated images exhibit the desired semantics.} The translation from $\bm{w}$ to $\bm{w}^*$ should be semantically correct;
    \item \textit{The source content is preserved.} A translated image $\bm{x}^* = G(\bm{w}^*)$ should maintain the attribute-invariant characteristics of the image and change only the attribute-specific properties of the source image $\bm{x}$.
    For example, in $\text{smile} \leftrightarrow \text{non-smile}$ translations, $\bm{x}^*$ should maintain the pose and identity of the source image $\bm{x}$;
    \item \textit{The output is multi-modal.} Translations are inherently ambiguous, thus it is desirable to have a model that generates different, plausible translations. Note that we are not here interested in images whose appearance is almost identical but e.g. individual hairs are placed very differently (such in StyleGAN~\cite{karras2019style});
    \item \textit{Training is multi-domain.} A single function $\mathcal{M}$ should be able to map each style code into multiple, different domains.
\end{enumerate}}

Simultaneously meeting these properties allows to overcome several limits of existing approaches and enable MMUITs using \emph{pre-trained} and \emph{frozen} unconditional GANs.

To fulfill the above properties, we propose a multilayer perceptron (MLP) as Implicit Style Function $\mathcal{M}$, which can be formulated as:
\begin{equation}
    \bm{w}^* = \mathcal{M}(\bm{w}, \bm{z}, \bm{d}),
\end{equation}
where $\mathcal{M}(\cdot)$ predicts a target latent code $\bm{w}^*$ from a source latent code $\bm{w}$, a noise vector $\bm{z}$ randomly sampled from a standard Gaussian (i.e., $\bm{z}\sim \mathcal{N}(\mathbf{0}, \mathbf{I}_n)$), and a target semantic vector $\bm{d}$. The source latent code $\bm{w}$ represents the latent code to be manipulated and $\bm{z}$ ensures the stochasticity of the manipulation process.

In order to learn such $\mathcal{M}$ in an unsupervised scenario, we assume to have a discriminator at training time which consists of two sub-tasks: (1) a binary classifier to distinguish the generated images from realistic images, and (2) a 
multi-label classifier $f: \bm{\mathcal{X}} \rightarrow \bm{\mathcal{D}}$ to distinguish whether the images are with expected visual semantics. Thus, the $\mathcal{M}$ is learnt in an adversarial approach.

\subsection{Learning the Implicit Style Function}
\label{sec:isf}
We propose to train our ISF $\mathcal{M}$ in an adversarial fashion~\cite{goodfellow2014generative} with the help of a multi-task discriminator $D$, which is learned together with $\mathcal{M}$. 
At training time, we only optimize the parameters in the networks $\mathcal{M}$ and $D$, while we do not update the parameters of the pre-trained generator $G$. Such a strategy helps our method requiring fewer resources than traditional models that train a generator from scratch.
The detailed architecture is depicted in \Cref{Fig:framework}.

We use several objective functions to constraint the training process and learn the ISF $\mathcal{M}$. 

\noindent\textbf{Realism \& Semantic Correctness.}  Inspired by recent literature (e.g.,~\cite{choi2018stargan,Liu_2021_CVPR}), we use a discriminator $D$ with two branches namely $D_{\text{cls}}$, devoted to domain classification, and $D_{\text{r/f}}$, which learns a binary classification determining whether an image $\bm{x}$ is real or fake. More formally, we ensure the \textit{realism} of the synthesized images through:
\begin{equation}
\begin{aligned}
    \mathcal{L}_{\text{r/f}} = \mathbb{E}_{\bm{w}, \bm{z}, \bm{d}}[ & \log D_{\text{r/f}}(G(\bm{w})) + \\ & \log(1-D_{\text{r/f}}(G(\mathcal{M}(\bm{w}, \bm{z}, \bm{d}))))].
\end{aligned}
\end{equation}
To ensure a manipulated code $\bm{w}^* = \mathcal{M}(\bm{w}, \bm{z}, \bm{d})$ generates an image with the specified semantics $\bm{d}$, we impose a domain classification loss when optimizing $D$ and $\mathcal{M}$. More formally, $D$ calculates the cross-entropy loss on the generated images over the non-manipulated latent codes by:
\begin{equation}
    \mathcal{L}^D_{\text{cls}} = \mathbb{E}_{\bm{w}, \bm{d}_0}\left[-\log D_{\text{cls}}(\bm{d}_{0}|G(\bm{w}))\right],
\end{equation}
where $\bm{d}_0 = f(G(\bm{w}))$ refers to the attribute vector extracted from the image generated from $G(\bm{w})$ by utilizing the pre-trained attribute classifiers $f$. We will explain $f$ in the Experiments Section. 
To learn classifying images, $\mathcal{M}$ tries to minimize the classification error of images generated with manipulated latent codes with:
\begin{equation}
    \mathcal{L}^{\mathcal{M}}_{\text{cls}} = \mathbb{E}_{\bm{s}, \bm{z}, \bm{d}}\left[-\log D_{\text{cls}}(\bm{d}|G(\mathcal{M}(\bm{s}, \bm{z}, \bm{d})))\right].
\end{equation}

\noindent\textbf{Content Preservation.} We propose to preserve the \emph{content} during the latent code manipulation with the following loss: 
\begin{equation}
\label{eq:lpips}
    \mathcal{L}_{\text{cont}} = \mathbb{E}_{\bm{s}, \bm{z}, \bm{d}}\left[\psi(G(\bm{s}), G(\mathcal{M}(\bm{s}, \bm{z}, \bm{d})))\right],
\end{equation}
where $\psi(\bm{x}_1, \bm{x}_2)$ estimates the perceptual distance  between image $\bm{x}_1$ and image $\bm{x}_2$ using an externally pre-trained network. 
Eq.~(\ref{eq:lpips}) minimizes the perceptual distance between $G(\bm{w})$ and any image generated with a manipulated code (i.e., $G(\mathcal{M}(\bm{w}, \bm{z}, \bm{d}))$). Similarly to Liu \emph{et al.}~\cite{Liu_2021_CVPR}, Eq.~(\ref{eq:lpips}) implies that some perceptual features should be maintained during the manipulation process. 
Although different perceptual distances can be used (e.g., the $\ell_2$ distance on VGG features~\cite{johnson2016perceptual}),  we implement $\psi(\bm{x}_1, \bm{x}_2)$ using the Learned Perceptual Image Patch Similarity (LPIPS) metric~\cite{zhang2018unreasonable}, which has been proved to be better aligned with the human perceptual similarity.

To further help the network to preserve the content of the input image, we also introduce a \textit{neighbouring constraint}:
\begin{equation}
\label{eq:nb}
    \mathcal{L}_{\text{nb}} = \mathbb{E}_{\bm{w}, \bm{z}, \bm{d}} \left[ \|\bm{w} - \mathcal{M}(\bm{w}, \bm{z}, \bm{d})\|_2 \right].
\end{equation}
Eq.~(\ref{eq:nb}) is motivated by previous literature that has shown neighbouring latent codes in StyleGAN exhibit similar semantic properties~\cite{karras2019style,karras2020analyzing}.

Finally, we also encourage the \textit{cycle consistency}~\cite{zhu2017unpaired,choi2018stargan,choi2019stargan}, which is typically used in most MMUIT models, to stabilize the training process.
The \textit{cycle consistency} loss encourages the image generated by the original latent code $\bm{w}$ and the image synthesized by the latent code mapped back into the original domain $\mathcal{M}(\bm{w}^*, \bm{z}, \bm{d}_0)$ to be as similar as possible. More formally, the cycle consistency loss is:
\begin{equation}
\label{eq:cyc_pixelwise}
    \mathcal{L}_{\text{cyc}} = \mathbb{E}_{ \bm{w},\bm{z}, \bm{d}, \bm{d}_0}\left[\|G(\bm{w}) - G(\mathcal{M}(\bm{w}^*, \bm{z}, \bm{d}_0))\|_1\right].
\end{equation}

\noindent\textbf{Multi-modality.} We aim at having multi-modal outputs through the randomly sampled noise $\bm{z}$ injected in $\mathcal{M}$. However, to further encourage $\mathcal{M}$ to produce diverse outputs, we employ the \textit{diversity sensitive} loss~\cite{choi2019stargan, mao2019mode}:
\begin{equation}
\label{eq:ds}
\begin{aligned}
        \mathcal{L}_{\text{ds}} = & \mathbb{E}_{\bm{w}, \bm{d},  (\bm{z}_1, \bm{z}_2)\sim{\mathcal{N}}(\mathbf{0}, \mathbf{I}_n)} [ \\ & \quad \| G(\mathcal{M}(\bm{w}, \bm{z}_1, \bm{d}) - G(\mathcal{M}(\bm{w}, \bm{z}_2, \bm{d}) \|_1],
\end{aligned}
\end{equation}
where $\mathcal{M}$ is conditioned on two random noise vectors $\bm{z}_1$ and $\bm{z}_2$. We maximize $\mathcal{L}_{\text{ds}}$ to have generated images that are as different as possible from each-other. To stabilize the learning process, we linearly decay the weight of the loss to zero during the training.

\noindent\textbf{Overall Objective}
\change{The full objective functions can be summarized as follow:
\begin{equation}
\begin{aligned}
    \min_{\mathcal{M}} \max_{D} = &  \lambda_{\text{r/f}}\mathcal{L}_{\text{r/f}} + \lambda_{\text{cls}}\mathcal{L}_{\text{cls}} + \lambda_{\text{cont}}\mathcal{L}_{\text{cont}} + \lambda_{\text{nb}}\mathcal{L}_{nb} + \\
    & 
    \lambda_{\text{cyc}}\mathcal{L}_{\text{cyc}} - \lambda_{\text{ds}}\mathcal{L}_{\text{ds}}
\end{aligned}
\end{equation}
where $\lambda_{\text{r/f}}$, $\lambda_{\text{cls}}$, $\lambda_{\text{cont}}$, $\lambda_{\text{nb}}$, $\lambda_{\text{cyc}}$ and $\lambda_{\text{ds}}$ are the hyper-parameters for each loss term. We note that the minus term before $\lambda_{ds}$ allows to maximize \Cref{eq:ds}.}
 
\subsection{Injecting the domain and multi-modality}
\label{sec:architecture}

\change{As shown in \Cref{Fig:framework}, $\mathcal{M}$ is implemented through two MLP function $f_1$ and $f_2$ and an AdaLN layer between them, which we explain in this Section. The non-linear $f_1$ function maps the concatenation of a domain vector and a random noise to a hidden latent variable $\bm{h} = f_1(\bm{d}\oplus\bm{z})$. 
Then, $\bm{h}$ goes through AdaLN, which is based on the widely used Adaptive Instance Normalization (AdaIN)~\cite{huang2017arbitrary}. 
Several state of the art approaches in I2I translation use AdaIN to inject the style features into the generating process and thus transfer the desired style into the output image. AdaIN assumes that images with the same style have a common mean and variance. Thus, existing methods usually extract the channel-wise mean and variance from images.
However, previous literature has shown that the channels in the  $\mathcal{W}^+$ latent codes of StyleGAN are correlated. 
For example, Karras \emph{et al.}~\cite{karras2019style} has shown that bottom style layers in $W$ (i.e., the first channels of $\mathcal{W}^+$) control high-level aspects such as pose, general hair style, face shape, and eyeglasses, middle-layers are about facial features, hair style, eyes open/closed, while higher layers are about color scheme and micro-structure.}

\change{Thus, we propose a simple modification to AdaIN by replacing Instance Normalization~\cite{ulyanov2016instance} with Layer Normalization: 
\begin{equation}
    \text{AdaLN}(\bm{w}, \bm{\gamma}, \bm{\beta}) = \bm{\gamma}\frac{\bm{w} - \mu(\bm{w})}{\sigma(\bm{w})} + \bm{\beta},
\end{equation}
where the $\mu(\cdot)$ and $\sigma(\cdot)$ are calculated by the Layer Normalization, $\bm{\gamma}$ and $\bm{\beta}$ are parameters generated by a linear mapping function with the input $\bm{h}$. 
Layer Normalization~\cite{ba2016layer} computes the normalization for each sample across all the channels, avoiding destroying valuable information in the latent code.
The parameters in the AdaLN module are learnt during the training. }

\change{The output of AdaLN goes through a second MLP function $f_2$ to obtain the edited style code $\bm{w}^* = f_2(\text{AdaLN}(\bm{w}, \bm{\gamma}, \bm{\beta}))$.}

\section{Experiments}
\label{sec:experiments}

We focus the experiments of the proposed ISF-GAN on StyleGAN v1~\cite{karras2019style} and StyleGAN v2~\cite{karras2020analyzing}, the two most prominent state-of-the-art unconditional GANs.
However, we note that our methodology can be applied to several GANs such as PG-GAN~\cite{karras2018progressive}.

\noindent\textbf{Baselines.} We compare our proposal with two state-of-the-art methods employing pre-trained and fixed unconditional GAN to manipulate faces: InterFaceGAN~\cite{shen2020interpreting} and StyleFlow~\cite{abdal2020styleflow}. 
Moreover, we compare the ISF-GAN to StarGAN v2~\cite{choi2019stargan} and its smooth extension~\cite{Liu_2021_CVPR} on gender translation, which are two state-of-the-art methods for MMUITs. \change{We use the officially released code for all the baselines in all these comparisons.} 

\noindent\textbf{Datasets.} InterFaceGAN~\cite{shen2020interpreting} and StyleFlow~\cite{abdal2020styleflow} were trained and tested on different conditions.
To have a fair evaluation between the proposed method and the baselines, we collect the following datasets:
\begin{itemize}
    \item \textit{Set$_1$}: we randomly sample 90K latent codes and collect the corresponding images through StyleGAN v1~\cite{karras2019style} pre-trained on FFHQ dataset~\cite{karras2019style}. 
    Then, we use an \emph{off-the-shelf} classifier~\cite{karras2019style} to label each latent code with the corresponding semantic attributes. We randomly split the dataset into 80K and 10K samples for the training and testing sets;
    \item \textit{Set$_2$}: we collect the dataset released by StyleFlow~\cite{abdal2020styleflow}, where 10K training images and 1K testing images are provided with annotated attributes and latent codes for StyleGAN v2~\cite{karras2020analyzing} pretrained on FFHQ dataset~\cite{karras2019style}.
\end{itemize}
In each dataset, we model four key facial attributes commonly-used by InterFaceGAN~\cite{shen2020interpreting} and StyleFlow~\cite{abdal2020styleflow} for analysis, including \textit{gender}, \textit{smile} (expression), \textit{age}, and \textit{eyeglasses}. We note that it is easily possible to plug in more semantic attributes as long as an attribute classifier is available. 

\noindent\textbf{Evaluation metrics.} We evaluate both the visual quality and the diversity of generated images using Fr\'echet Inception Distance (FID)~\cite{martin2017fid}, the Learned Perceptual Image Patch Similarity (LPIPS)~\cite{zhang2018unreasonable} and the Accuracy, evaluated through an \emph{off-the-shelf} classifier provided by~\cite{karras2019style}. The details on evaluation metrics and protocols are further described in the Supplementary Materials (SI).

Moreover, we propose to use the state-of-the-art face recognition method ArcFace~\cite{deng2019arcface} to evaluate the content preservation in MMUITs. We define a new metric called Face Recognition Similarity (FRS) that estimates the similarity between features extracted from two facial images. 
More formally: 
    \begin{equation}
        \text{FRS} = \mathbb{E}_{\bm{s}, \bm{z}, \bm{d}} \left[ \langle \omega(G(\bm{s})), \omega(G(\mathcal{M}(\bm{s}, \bm{z}, \bm{d})))\rangle \right]
    \end{equation}
    where $\omega(\cdot)$ refers to the pre-trained ArcFace network, $\langle\cdot, \cdot\rangle$ refers to cosine similarity between two input vectors.
    The extracted face regions are resized to 112$\times$112 before being fed into $\omega$. 
    An high FRS score shows the two faces have more similar identities, which indicates that a model better preserves the identities of faces during image manipulations.

\noindent\textbf{Training Details.}
ISF-GAN is trained for 40K iterations with batch size 4.
The training time takes about 16 hours on 4 Tesla V100 GPUs with our implementation in PyTorch~\cite{paszke2017automatic}. Compared to the state-of-the-art MMUIT model StarGAN v2~\cite{choi2019stargan}, the training time is reduced by half while having a much higher resolution (i.e., from $256 \times 256$ to $1024 \times 1024$). 
\change{We set $\lambda_{\text{r/f}}=1$, $\lambda_{\text{cls}}=1$ $\lambda_{\text{cont}}=1$, $\lambda_{\text{nb}}=0.1$, $\lambda_{\text{cyc}}=1$ and $\lambda_{\text{ds}}=2.0$. }
We adopt the non-saturating adversarial loss~\cite{goodfellow2014generative} with R1 regularization~\cite{mescheder2018training} using $\gamma$=1. We use the Adam~\cite{kingma2015adam} optimizer with $\beta_1$=0 and $\beta_2$=0.99. The learning rates for $\mathcal{M}$ and $D$ are set to $10^{-5}$.

\begin{figure}[h]	
	\renewcommand{\tabcolsep}{1pt}
	\centering
	\begin{tabular}{c}
	    \includegraphics[width=0.95\linewidth]{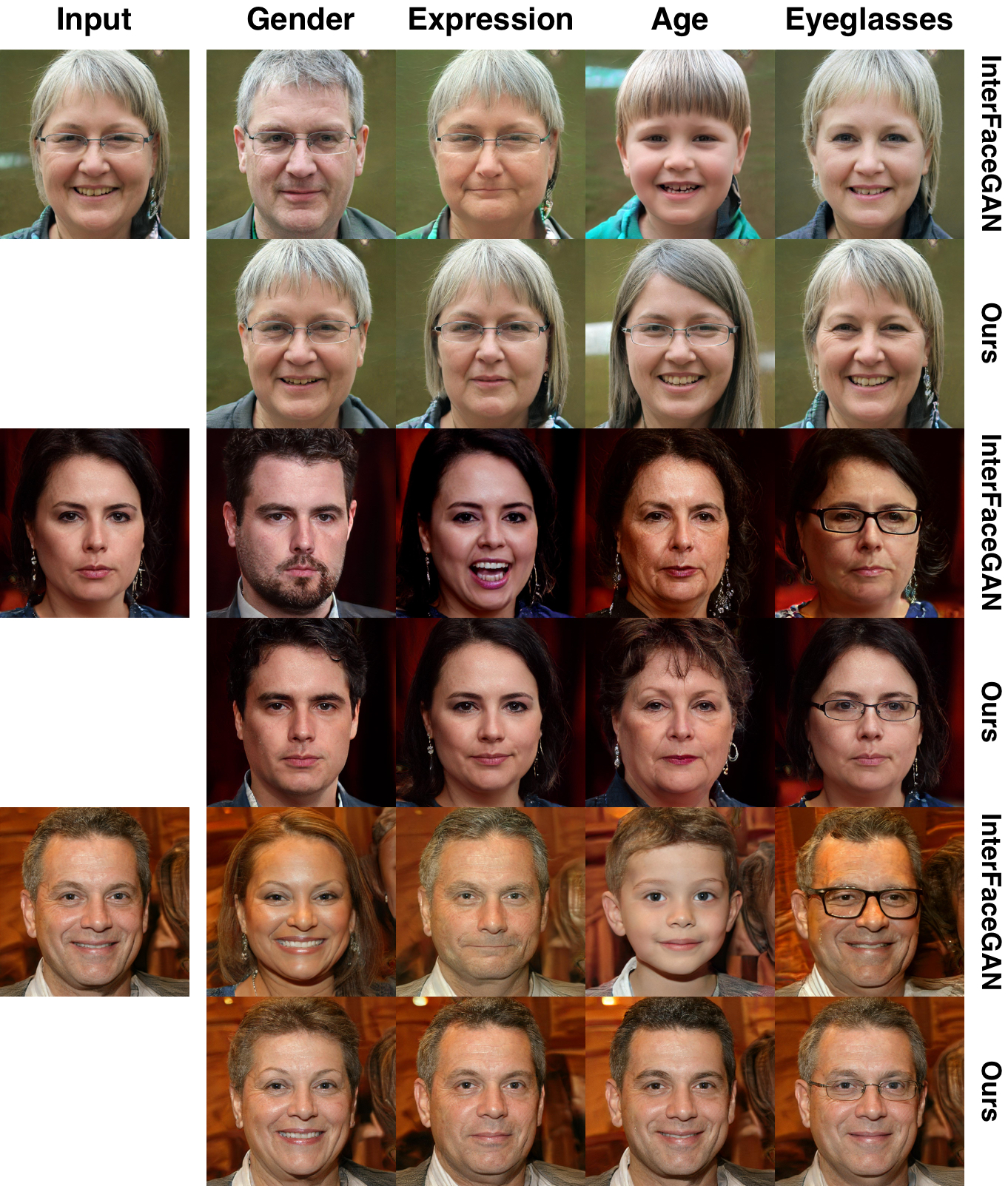}
	\end{tabular}
	\caption{
	\change{Visual comparisons between InterFaceGAN~\cite{shen2020interpreting} and ISF-GAN on various attribute manipulations tested on the \textit{Set}$_1$. }
	} 
	\label{Fig:visual-comparison-set1}
\end{figure}

\begin{figure}[h]	
	\renewcommand{\tabcolsep}{1pt}
	\centering
	\begin{tabular}{c}
	    \includegraphics[width=0.95\linewidth]{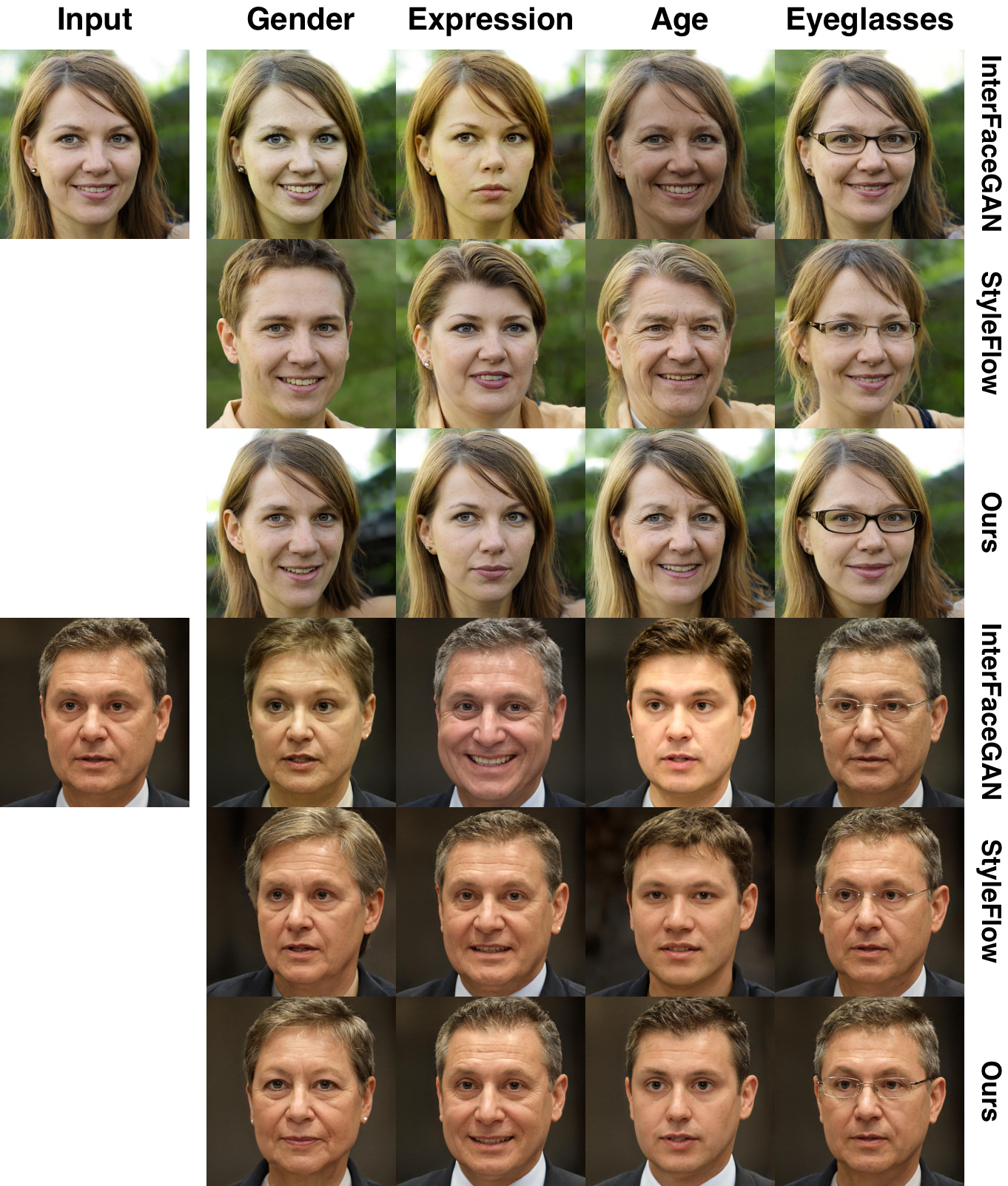}
	\end{tabular}
	\caption{
	\change{Visual comparisons between InterFaceGAN~\cite{shen2020interpreting}, StyleFlow~\cite{abdal2020styleflow} and our ISF-GAN on the \textit{Set}$_2$.
	}} 
	\label{Fig:visual-comparison-set2}
\end{figure}

\subsection{Latent Codes Manipulation}
We start by evaluating the performance of our framework on manipulating StyleGAN latent codes. Given a randomly sampled StyleGAN latent code $\bm{w}$ with its corresponding attribute vector $\bm{d}$, we choose a desire attribute vector $\bm{d}^*$ and use $\mathcal{M}$ to estimate a target $\bm{w}^*$. Intuitively, an ideal model should well manipulate the original latent code $\bm{w}$ to make the generated image  $\bm{x}^* = G(\bm{w}^*)$ with correct semantics. 
\Cref{Fig:visual-comparison-set1} suggests that the proposed ISF-GAN correctly learns the different semantics in StyleGAN v1 latent codes and allows to generate realistic translations. 
Compared to InterFaceGAN, our synthesized faces preserve the identity of the input image better, while correctly lying on the desired semantics. Similarly, as shown in \Cref{Fig:visual-comparison-set2}, the proposed method not only translate the input images with correct visual semantics but better preserves the face identity also in StyleGAN v2. 

As shown in \Cref{tab:main-results}, the overall quantitative evaluation verifies the superiority of the proposed method. The proposed ISF-GAN achieves better performance on FID and LPIPS, which indicates ISF-GAN synthesizes more realistic images and enables the generation of diverse images. This observation is consistent with the visual comparisons presented in \Cref{Fig:visual-comparison-set1} and \Cref{Fig:visual-comparison-set2}. 
\Cref{tab:main-results} shows that InterFaceGAN scores 0.00 in the LPIPS as the model is deterministic and does not allow diverse translations for the same input code. The only way to have diversity is to generate style latent codes in very close range to the translated latent code. This way scores 0.06 and 0.05 for $Set_1$ and $Set_2$, which is a result comparable to StyleFlow. However, changing the style codes in this way might affect the entire content of the generated image, including the identity of the person. Thus, it is not trivial to have multi-modal translations with InterFaceGAN.

\begin{figure}[!ht]	
	\renewcommand{\tabcolsep}{1pt}
	\centering
	\begin{tabular}{c}
	    \includegraphics[width=0.95\linewidth]{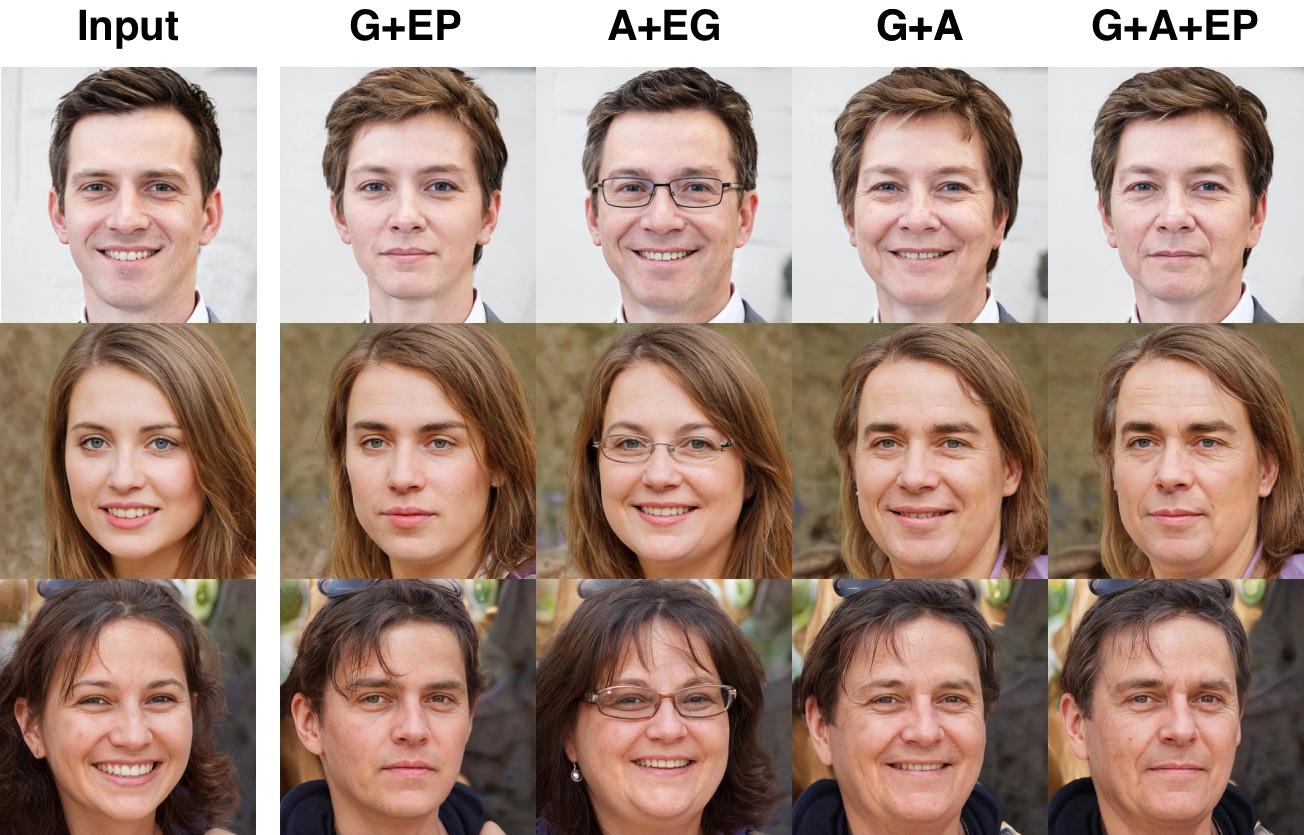}
	\end{tabular}
	\caption{
	\change{Visual results on multi-attributes manipulation at a time, where ``G", ``EP", ``A" and ``EG" refer to \textit{gender}, \textit{expression}, \textit{age} and \textit{eyeglasses} attributes, respectively. \vspace{-1em}}
	} 
	\label{Fig:multi-attributes}
\end{figure}

We note that ISF-GAN can modify more than one attribute at a time, as shown in \Cref{Fig:multi-attributes} while InterFaceGAN~\cite{shen2020interpreting, shen2020interfacegan} relies on multiple learned hyper-planes and requires users to translate multiple attributes in multiple steps (e.g. female$\rightarrow$male$\rightarrow$smile). 
StyleFlow~\cite{abdal2020styleflow} can manipulate multiple attributes at a time, but achieves lower accuracy than our model on the domain label used as a target of the translation (see mAcc in \Cref{tab:main-results}).

We also note that we could not test Style-Flow on $Set_1$ as the authors did not release the model for StyleGAN v1. Thus, we could not have a fair comparison with them. 

\begin{table*}[!ht]
    \centering
    \small
    \begin{tabularx}{0.78\linewidth}{l rrrr rrrr}
    \toprule
    \multirow{2}{*}{\textbf{Model}} & \multicolumn{4}{c}{\textit{Set}$_1$} & \multicolumn{4}{c}{\textit{Set}$_2$}\\ \cmidrule(lr){2-5}  \cmidrule(lr){6-9} 
    & \textbf{FID}$\downarrow$ & \textbf{LPIPS}$\uparrow$ &
    \textbf{FRS}$\uparrow$ & \textbf{mAcc}(\%)$\uparrow$ 
    & \textbf{FID}$\downarrow$ & \textbf{LPIPS}$\uparrow$ &
    \textbf{FRS}$\uparrow$ & \textbf{mAcc}(\%)$\uparrow$ \\ 
    \midrule
    StyleFlow~\cite{abdal2020styleflow} & - & - & - & - & 50.42 & 0.04 & 0.62 & 72.23\\ %
    InterFaceGAN~\cite{shen2020interfacegan} & 50.89 & 0.00 & 0.32 & 94.17 & 29.85 & 0.00 & \textbf{0.66} & 78.95 \\
    Ours & \textbf{29.44} & \textbf{0.16} & \textbf{0.75} & \textbf{95.45} & \textbf{23.93} & \textbf{0.22} & \textbf{0.66} & \textbf{96.52} \\
    \bottomrule
    \end{tabularx}
    \caption{\change{Quantitative comparisons on image quality, diversity, content preservation and accuracy of generated images based on pre-trained StyleGAN v1~\cite{karras2019style}/v2\cite{karras2020analyzing}. The proposed ISF-GAN outperforms all state-of-the-art methods.}
    }
    \label{tab:main-results}
\end{table*}

\begin{figure}[!ht]	
	\renewcommand{\tabcolsep}{1pt}
	\centering
	\begin{tabular}{c}
	    \includegraphics[width=0.92\linewidth]{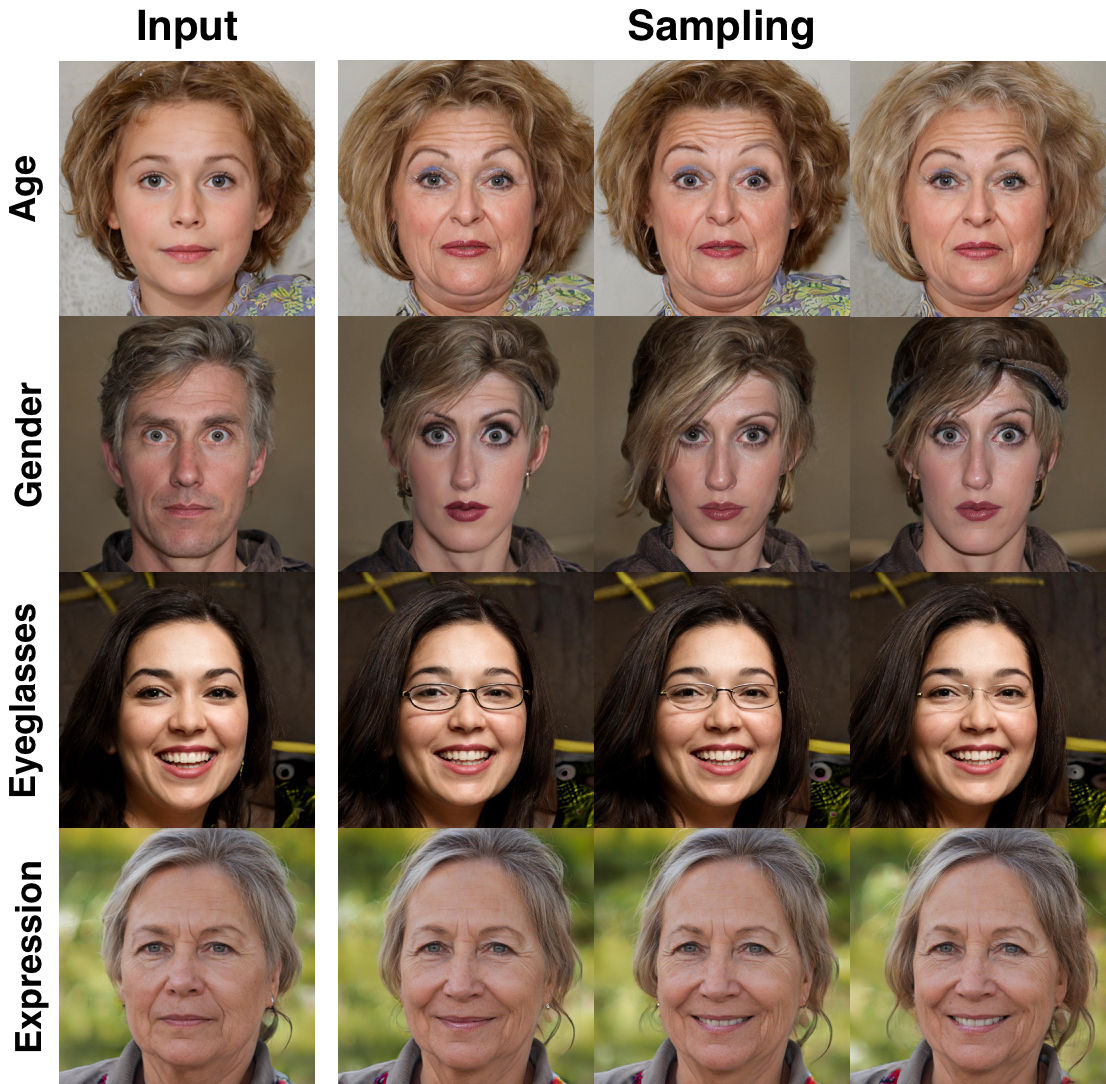}
	\end{tabular}
	\caption{\change{Multi-modal results obtained randomly sampling multiple $\bm{z}$ and collect diverse synthesized images.\vspace{-1em} }
	} 
	\label{Fig:sampling}
\end{figure}

\vspace{0.2cm}
\noindent\textbf{Content preservation.} Quantitatively evaluating content preservation is very challenging. However, in face translation, the content is mainly related to the identity of people. Thus, we here evaluate whether the identity is preserved during the manipulation tasks. \Cref{tab:main-results} shows that FRS increases by 121\% in InterFaceGAN with $Set_1$. In $Set_2$, improvements are only slight perhaps due to the smoother style latent space of StyleGANv2. 
\Cref{Fig:sampling} shows that the identity preservation is also stable when we use randomly sample different noise codes $\bm{z}$.

\vspace{0.2cm}
\noindent\textbf{Latent space separation.} 
The FID, LPIPS and FRS results do not come at the expense of the semantic correctness of generated images.
We calculate the classification accuracy of the generated images by using off-the-shelf classifiers, in which the test set is composed of 1,000 positive samples and 1,000 negative samples for each manipulated attribute. In \Cref{tab:main-results}, our proposed method achieves better performance in both the \textit{Set}$_1$ and \textit{Set}$_2$. 
We also compare the semantic accuracy of our method with the baselines in SI Table I. Our proposed method achieves the best performances on almost all of the four attributes.
It indicates that ISF-GAN correctly disentangles the different semantics contained in the StyleGAN latent code better than the start-of-the-art methods, thus allowing to synthesize images into the desired semantic specified by attribute vector $\bm{d}$.

\begin{figure*}[!ht]	
	\renewcommand{\tabcolsep}{1pt}
	\centering
	\begin{tabular}{c}
	    \includegraphics[width=0.9\linewidth]{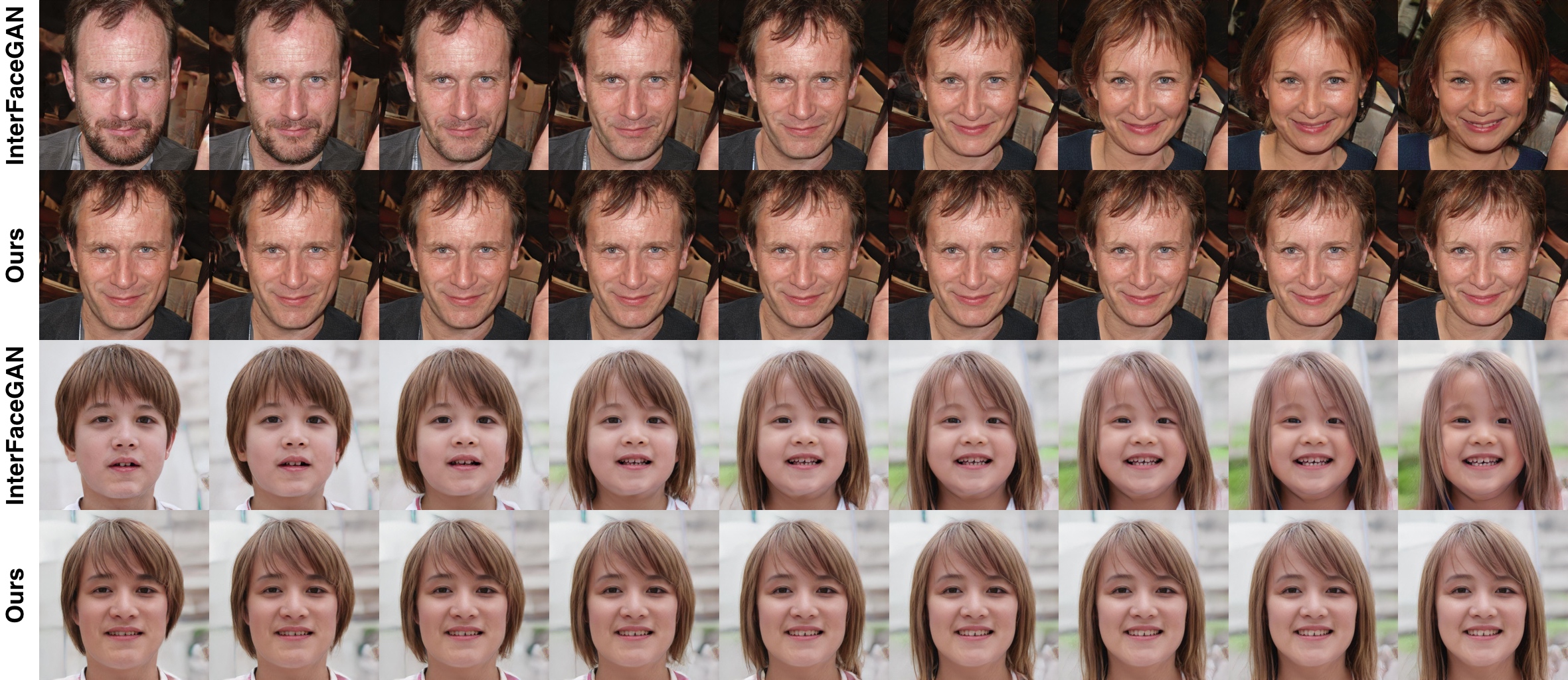}
	\end{tabular}
	\caption{\change{Comparisons on gender interpolations between InterFaceGAN~\cite{shen2020interpreting} and our proposed method tested on \textit{Set}$_1$. Compared to InterFaceGAN, our method preserves better the face identity along the interpolations.}} 
	\label{Fig:interpolation-set1}
\end{figure*}

\subsection{Semantic Interpolation}
Evaluating semantic interpolations is important for two main reasons. On the one hand, semantic interpolation allows to generate a smooth flow between two semantics (e.g., \emph{males} and \emph{females}), enabling users to generate intermediate images (e.g., non-binary gender or people transitioning gender) that are rarely present in existing datasets. On the other hand, interpolating latent codes allows qualitatively and quantitatively evaluating the smoothness of the latent space.

To semantically interpolate in the StyleGAN space, we first randomly sample a style $\bm{s}$, then we manipulate its semantic with $\bm{s}^* = \mathcal{M}(\bm{s}, \bm{z}, \bm{d})$, where $\bm{d}$ changes exactly one attribute from the attribute vector of $\bm{s}$. Then, we linearly interpolate $T$ equi-distant points along the interpolation line between $\bm{s}$ and $\bm{s}^*$. Finally, we generate the images in the path: $\left[G(\bm{s}_0), G(\bm{s}_1), \ldots, G(\bm{s}_T)) \right]$, where $\bm{s}_0 = \bm{s}$ and $\bm{s}_T = G(\bm{s}^*)$.

SI Fig. 15 shows multiple semantic interpolations where we change the age, gender, expression and eyeglasses of some randomly sampled StyleGAN latent codes. We observe that the interpolations are very smooth, having almost imperceptible perceptual differences between neighbouring images.

To quantitatively evaluate the smoothness of interpolated images we rely on the Perceptual Path Length (PPL)~\cite{karras2019style} and a metric we propose: the Perceptual Interpolation Range (PIR).
PPL measures the perceptual variation between pairs of images generated under small perturbations $\epsilon$ in the latent space:
\begin{equation}
    PPL = \mathbb{E}_{\bm{s}, \bm{s}^*\sim\mathcal{S}, t \sim U(0,1)} \left[ \frac{1}{\epsilon^2} \Phi(\text{lerp}(\bm{s}, \bm{s}^*, t), \text{lerp}(\bm{s}, \bm{s}^*, t+ \epsilon)) \right]
\end{equation}
where $\Phi(\bm{s}_i, \bm{s}_j) = d(G(\bm{s}_i), G(\bm{s}_j))$, and $d(\cdot, \cdot)$ refers to the L2 distance between VGG~\cite{simonyan2014very} features.

\change{PIR instead measures the semantic smoothness along the entire interpolation path between two latent codes.
Specifically, PIR computes the range between the maximum and the minimum perceptual distance between two consecutive style codes along the interpolation path as $\max_{t=1}^T \phi(\bm{s}_{t-1}, \bm{s}_t) - \min_{t=1}^T \phi(\bm{s}_{t-1}, \bm{s}_t)$. 
Very smooth interpolations between $\bm{s}_0$ and $\bm{s}_T$ are expected to have $\max_{t=1}^T \phi(\bm{s}_{t-1}, \bm{s}_t) \approx \min_{t=1}^T \phi(\bm{s}_{t-1}, \bm{s}_t)$, thus resulting in $PIR \approx 0$. Interpolations with abrupt changes would have $\max_{t=1}^T \phi(\bm{s}_{t-1}, \bm{s}_t) \gg \min_{t=1}^T \phi(\bm{s}_{t-1}, \bm{s}_t)$, resulting in a high PIR.
Since the perceptual distance along the interpolation path might be different depending on the sampled $\bm{s}_0$ and $\bm{s}_T$ we normalize the computed range by $\phi(\bm{s}_1, \bm{s}_T)$. Thus:}
\begin{equation}
    PIR = \mathbb{E}_{\bm{s}_1, \bm{s}_T\sim\mathcal{S}} \left[\frac{\max_{t=1}^T \phi(\bm{s}_{t-1}, \bm{s}_t)- \min_{t=1}^T \phi(\bm{s}_{t-1}, \bm{s}_t)}{\phi(\bm{s}_1, \bm{s}_T) + \epsilon}\right]
\end{equation}
where $\epsilon$ is used for numerical stability. PIR uses the LPIPS~\cite{zhang2018unreasonable} as the perceptual distance $d$, as LPIPS has been shown to be well aligned with human perceptual similarity.

\Cref{tab:result-PIR} shows the quantitative results of the tested models. Both the PPL and PIR results show that the latent codes of our model are less entangled and that the perceptual path is substantially shorter than InterFaceGAN in \textit{Set}$_1$. Similarly, although the PPL results are comparable between StyleFlow, InterFaceGAN and the proposed ISF-GAN in \textit{Set}$_2$, PIR results suggest that there are less drastic changes between images generated along the interpolation path in ISF-GAN.

\begin{table}[!ht]
    \centering
    \small
    \begin{tabularx}{\columnwidth}{@{}X c rrrr rrrr@{}}
    \toprule
    \textbf{Method} & \textbf{Dataset} & \textbf{PPL}$\downarrow$ & \textbf{PIR}$\downarrow$ \\
    \midrule
        InterFaceGAN~\cite{shen2020interfacegan} & \multirow{2}{*}{\textit{Set}$_1$} & 28.76 & 0.06 \\
        Ours & & \textbf{10.23} & \textbf{0.01} \\ %
    \midrule
        StyleFlow~\cite{abdal2020styleflow} & \multirow{3}{*}{\textit{Set}$_2$} & 63.06 & 0.11\\
        InterFaceGAN~\cite{shen2020interfacegan} & & \textbf{60.47} & \textbf{0.07} \\ %
        Ours & & 62.06 & \textbf{0.07} \\
        \bottomrule
    \end{tabularx}
    \caption{\change{Comparisons on the smoothness of semantic interpolations on \textit{gender} translation.}}
    \label{tab:result-PIR}
    \vspace{-0.5em}
\end{table}

\begin{figure}[!ht]	
	\renewcommand{\tabcolsep}{1pt}
	\centering
	\begin{tabular}{c}
	    \includegraphics[width=\linewidth]{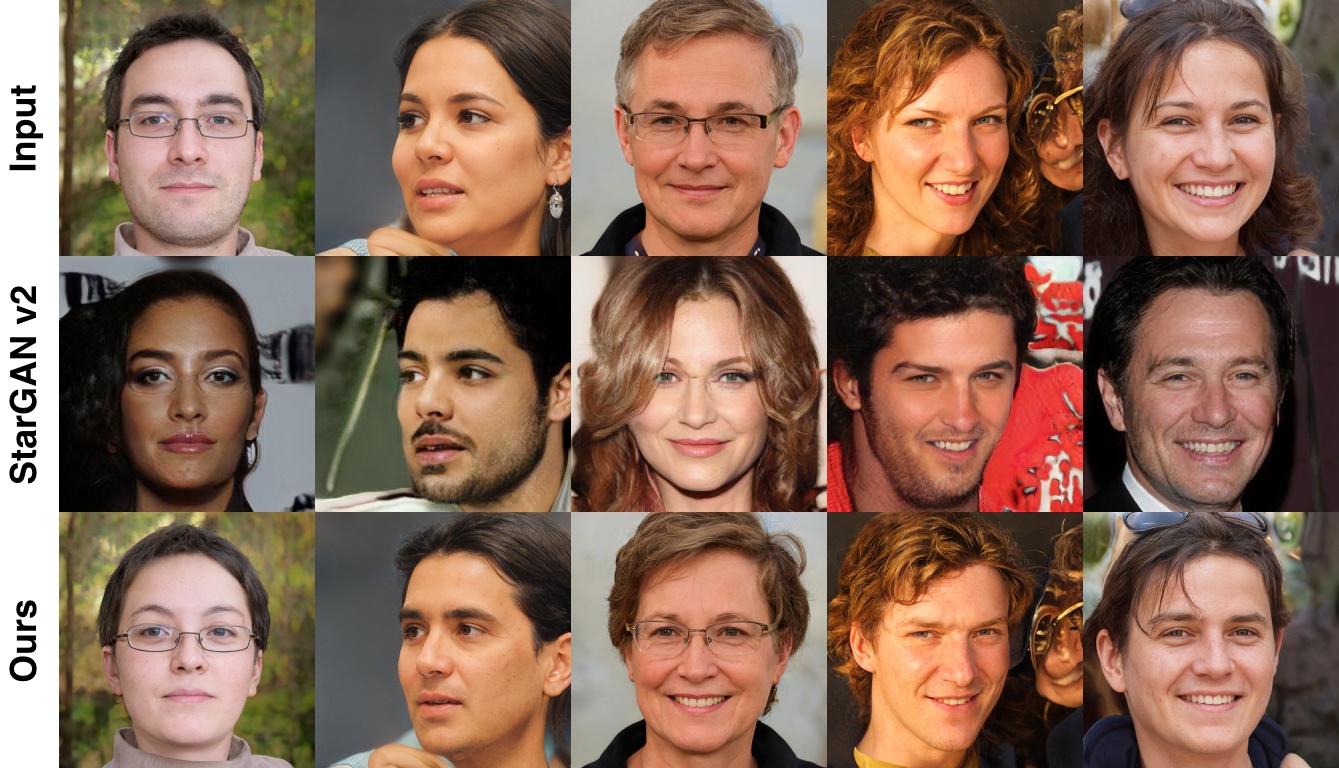}
	\end{tabular}
	\caption{\change{Visual comparisons between the proposed ISF-GAN and StarGAN v2~\cite{choi2019stargan} on gender translation.}
	\vspace{-0.5em}
	}
	\label{Fig:vs-starganv2}
\end{figure}
\begin{figure}[ht]	
	\renewcommand{\tabcolsep}{1pt}
	\centering
	\begin{tabular}{c}
	    \includegraphics[width=0.95\linewidth]{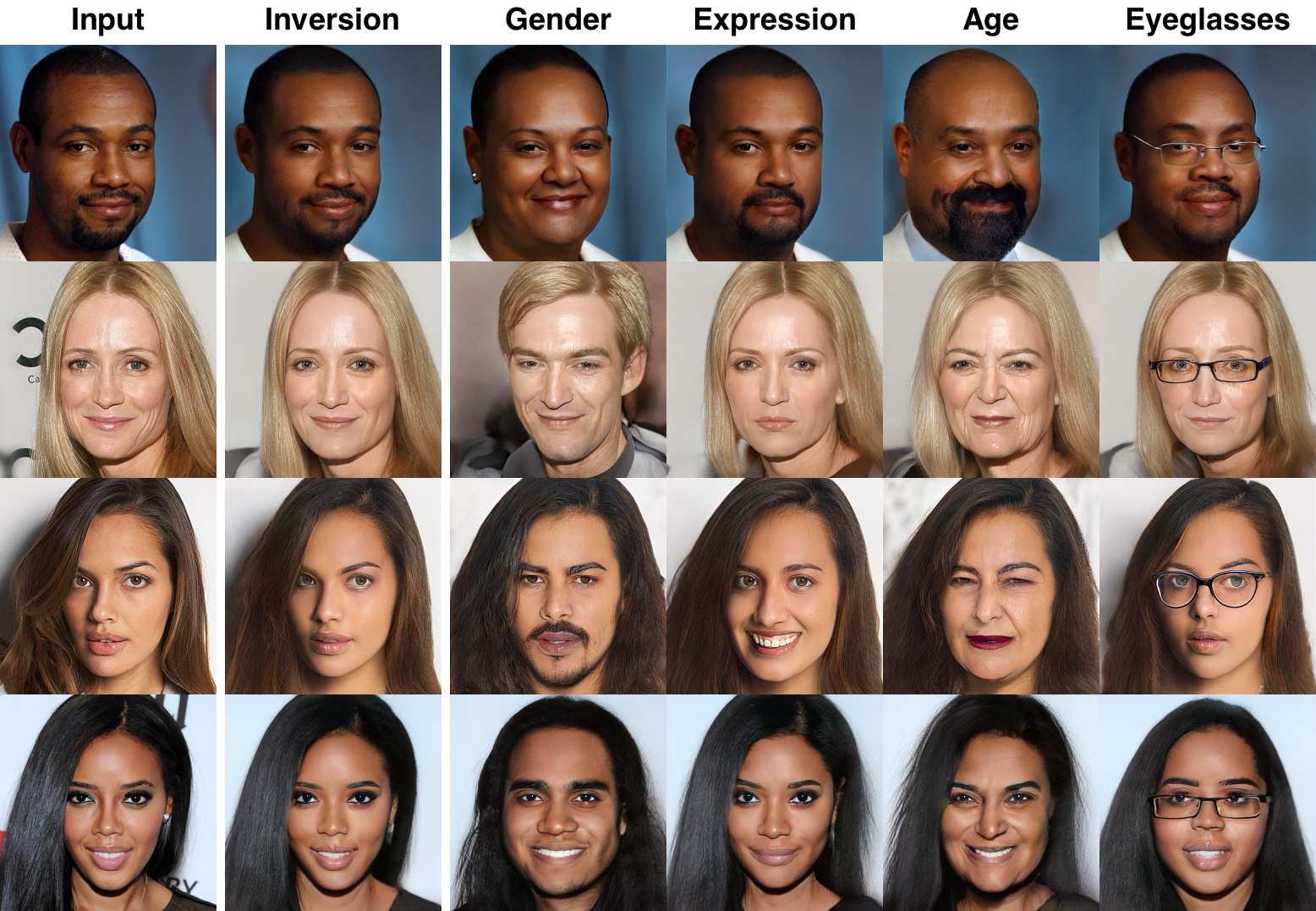}
	\end{tabular}
	\caption{\change{Manipulations of real images from CelebA-HQ~\cite{karras2018progressive} through ISF-GAN  and an image embedding method~\cite{abdal2019image2stylegan}.}
	}
	\label{Fig:real-image}
	\vspace{-0.5em}
\end{figure}

\subsection{Comparisons with Traditional MMUIT Models}

We evaluate ISF-GAN with the state-of-the-art model of MMUITs, namely StarGAN v2~\cite{choi2019stargan}. We test all the models on gender translation, following the main results of StarGAN v2.

ISF-GAN achieves better performance compared to StarGAN v2 (20.67 vs 71.20 on FID score).
\Cref{Fig:vs-starganv2} shows that ISF-GAN not only generates more realistic images with correct semantics but also preserves better face identity and background. We also compare our model with SmoothLatent~\cite{Liu_2021_CVPR}, which improves StarGAN v2 to have smooth inter-domain interpolations. ISF-GAN achieves better performance compared to SmoothLatent (0.050 vs 0.062 on PIR score). SI Fig. 13 shows that the inter-domain interpolations are smoother in both visual semantics and background.

We note that training ISF-GAN requires fewer resources than training StarGAN v2 and its extension.

\vspace{-1em}
\subsection{Manipulating Real Face Images}
\Cref{Fig:real-image} shows an example of manipulation of real images. First, we map the source images into the StyleGAN latent space through Abdal \emph{et al.}~\cite{abdal2019image2stylegan}, in which they search the latent code that best approximates the given image. Then, we use our model to translate images into multiple semantics. The results show that ISF-GAN can achieve high-quality results with such inverted latent codes of real images. Moreover, it indicates a promising application that ISF-GAN is used in conjunction with an \emph{off-the-shelf} method to do GAN inversion in StyleGAN (either through an encoder or an optimization method) for unsupervised image-to-image translation.
SI Fig. 11 shows some additional examples on CelebA-HQ.

\subsection{Beyond Face Translations}

To verify the generalization of the proposed method, we apply it to a two-domain image translation (e.g., Cat $\leftrightarrow$ Dog) task. We collect a pre-trained StyleGAN v2 on the AFHQ dataset~\cite{choi2019stargan} (using only ``cat" and ``dog" images). The model is trained with images at lower resolution (256$\times$256 resolution), batch size 8 and 1e5 iterations on 4 Tesla V100 GPUs. SI Fig. 12 shows several randomly sampled generated images based on the pre-trained StyleGAN v2.  

\begin{figure}[!ht]	
	\renewcommand{\tabcolsep}{1pt}
	\centering
	\begin{tabular}{c}
	    \includegraphics[width=\linewidth]{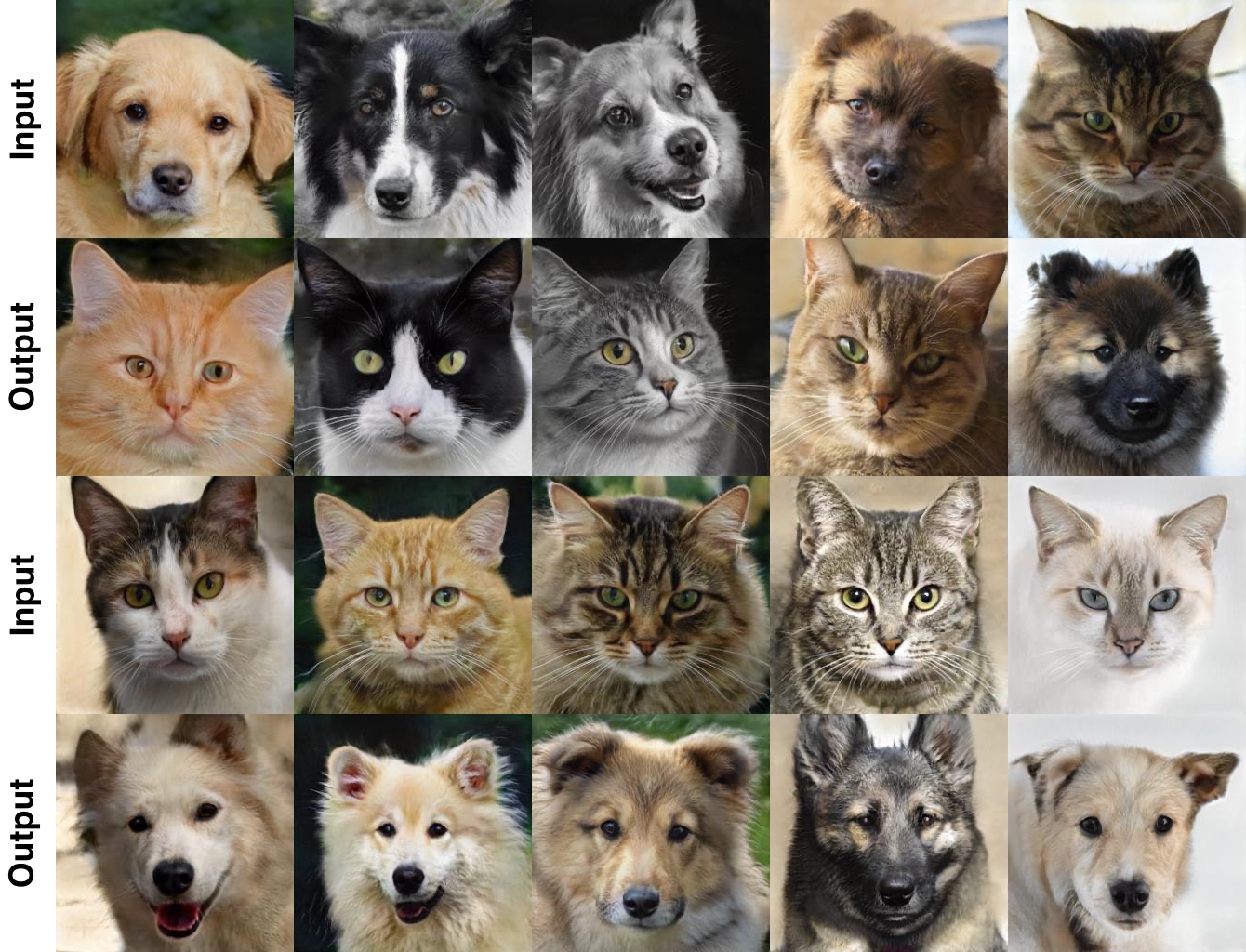}
	\end{tabular}
	\caption{Visual results of ISF-GAN for Cat $\leftrightarrow$ Dog translations based on a pre-trained StyleGAN v2. \vspace{-1em}
	}
	\label{Fig:stylegan2-cat2dog}
\end{figure}

Similar to the procedure in Section~\ref{sec:experiments}, we randomly sample 11K latent codes (10K for training and 1K for testing) in the $\mathcal{W}^+$ space and collect the corresponding images through the pre-trained GAN. Furthermore, in order to automatically distinguish the cat and dog faces, we also train a two-class classifier (i.e., a standard ResNet-50~\cite{he2016deep}) on the AFHQ dataset. Finally, we train the proposed ISF-GAN on such pre-trained GAN and aim to do Cat$\leftrightarrow$Dog translation. As shown in \Cref{Fig:stylegan2-cat2dog}, ISF-GAN performs well on this task, which indicates the excellent generalization ability of the proposed method.

\subsection{Ablation Study}
We here study the importance of each component of our proposed method. As shown in \Cref{tab:ablation}, the FID, LPIPS, FRS and PIR values for all the configurations where we removed or replaced components of our proposal \Cref{tab:ablation}-A. We repeated the experiments five times to compute the standard deviation of each metric.

First, we observe that FID and FRS significantly decrease by removing the neighbouring constraints $\mathcal{L}_{nb}$, but the diversity in the generated images (see \Cref{tab:ablation}-B) is evidently improved. This result suggests that constraining the latent codes to be similar in features to neighbouring latent codes helps disentangle the attributes from the attribute-invariant part of the image, thus improving the content preservation and quality of synthesized images.

Second, when we remove $\mathcal{L}_{cont}$, which encourages manipulated images to be perceptually similar to the original image, the LPIPS also improves but FID and FRS decrease. It shows a similar tendency as \Cref{tab:ablation}-B. However, $\mathcal{L}_{cont}$ has a more significant effect on the interpolation smoothness compared to $\mathcal{L}_{nb}$.  
This result suggests that perceptual loss is significant for better image quality and content preservation. From qualitative results, we see that $\mathcal{L}_{cont}$ helps the ISF-GAN better preserve some distinct face characteristics (e.g., the chin shape). The network without $\mathcal{L}_{cont}$ generates more diverse images because it also changes the identity of people.

We also ablate the proposed AdaLN by replacing it with AdaIN. \Cref{tab:ablation}-D shows that all metrics get worse. This suggests that StyleGAN latent codes do not have a channel-wise style distribution and that a channel-wise normalization destroys valuable information in it. Thus, the experiment verifies that the proposed AdaLN is more suitable for manipulating the latent codes of StyleGAN.

\begin{table}[!ht]
    \centering
    \small
    \begin{tabularx}{\columnwidth}{@{}X r rrrr@{}}
    \toprule
    \textbf{Model} & \textbf{FID}$\downarrow$ & \textbf{LPIPS}$\uparrow$ &
    \textbf{FRS}$\uparrow$ & \textbf{PIR}$\downarrow$ \\
    \midrule
    A: Our proposal & \textbf{29.44}$_{\pm.08}$ & .16$_{\pm.00}$ & \textbf{.754}$_{\pm.005}$ & \textbf{.009$_{\pm.003}$}  \\ %
    B: A w/o $\mathcal{L}_{nb}$ & 36.31$_{\pm.15}$ & \textbf{.29}$_{\pm.00}$ & .512$_{\pm.001}$ & .029$_{\pm.009}$ \\ %
    C: A w/o $\mathcal{L}_{cont}$ & 31.20$_{\pm.13}$ & .29$_{\pm.00}$ & .646$_{\pm.003}$ & .014$_{\pm.002}$ \\ %
    D: A w AdaIN & 30.44$_{\pm.13}$ & .13$_{\pm.00}$ & .695$_{\pm.002}$ & .012$_{\pm.003}$ \\ %
    \bottomrule
    \end{tabularx}
    \caption{\change{Ablation study on our proposed losses and AdaLN on gender manipulations with $Set_1$.}
    }
    \label{tab:ablation}
    \vspace{-0.5em}
\end{table}

\change{Finally, we ablate the value of $\lambda_{ds}$, which controls how much the model should focus on diversity. 
As expected, \Cref{tab:ablation-lambda} shows that the bigger is the value, the higher is the diversity of generated images. However, we also see that the changes on FID, FRS and PIR seem to be negatively correlated with the changes on LPIPS. 
This result is explained by the image quality and diversity trade-off previously discussed in \cite{Liu_2021_CVPR}. 
It is easier for the model to collapse to images very similar to the target domain, thus obtaining a lower FID and LPIPS, than creating very diverse images that sometimes contain features (e.g. a lot of different hairstyles) that are not present in the source domain images.}

\change{In our paper, we chose to have a higher FID but having higher LPIPS. However, we note that our improvements in FID from the state of the art models are still substantial, as depicted in \Cref{tab:main-results}.}

\begin{table}[!ht]
    \centering
    \small
    \begin{tabularx}{\columnwidth}{@{}X r rrrr@{}}
    \toprule
    \textbf{Model} & \textbf{FID}$\downarrow$ & \textbf{LPIPS}$\uparrow$ &
    \textbf{FRS}$\uparrow$ & \textbf{PIR}$\downarrow$ \\
    \midrule
    Our proposal w $\lambda_{ds}=0.2$ & 22.60 & 0.05 & 0.76 & 0.05 \\
    Our proposal w $\lambda_{ds}=1$ & 23.64 & 0.16 & 0.67 & 0.06 \\
    Our proposal w $\lambda_{ds}=2$ & 23.93 & 0.22 & 0.66 & 0.07 \\
    \bottomrule
    \end{tabularx}
    \caption{\change{Image quality - diversity trade-off on $Set_2$. 
    }}
    \label{tab:ablation-lambda}
    \vspace{-0.5em}
\end{table}

\vspace{-1em}
\section{Conclusion}
\label{sec:conclusion}

In this paper, we propose an Implicit Style Function to manipulate the semantics of StyleGAN latent codes. We show that our approach enables multi-modal manipulations controlled by an attribute vector through qualitative and quantitative results. Moreover, these manipulations are smoother than the state-of-the-art approaches and better preserve the attribute-invariant visual parts of the original latent code.
Altogether, our experimental results show that our model represents a competitive approach for applying \emph{pre-trained} and \emph{fixed} large-scale generators to Multi-modal and Multi-domain Unsupervised Image-to-image Translation tasks.

\ifCLASSOPTIONcaptionsoff
  \newpage
\fi

\footnotesize
\bibliographystyle{IEEEtran}
\bibliography{ref}

\vspace{-4em}
\begin{IEEEbiography}[{\includegraphics[width=0.8in,clip,keepaspectratio]{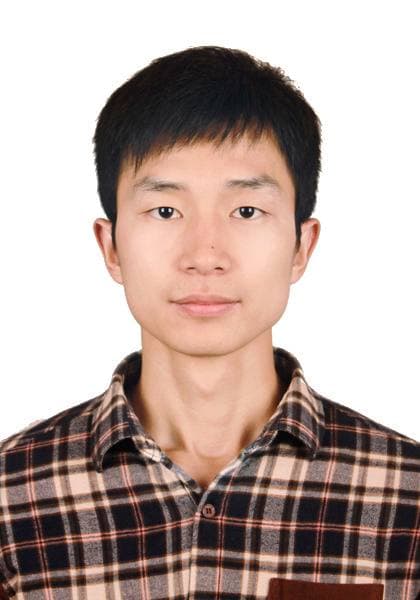}}]{Yahui Liu}
is a Ph.D. student in the Department of Information Engineering and Computer Science with the University of Trento, Italy. Before that, he received his B.S. degree and M.S. degree from Wuhan University, China, in 2015 and 2018, respectively. His major research interests are machine learning and computer vision, including unsupervised learning and image domain translation.
\end{IEEEbiography}
\vspace{-4em}

\begin{IEEEbiography}[{\includegraphics[width=0.8in,height=1.25in,clip,keepaspectratio]{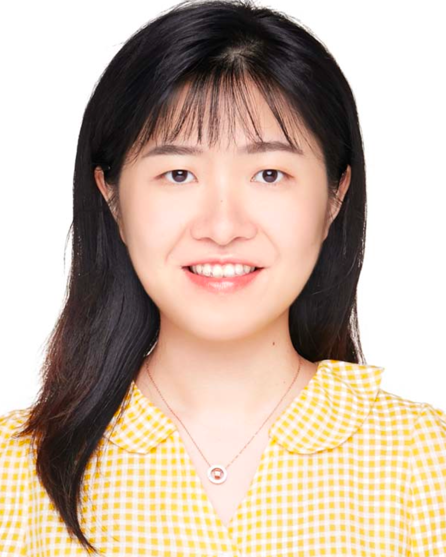}}]{Yajing Chen} received the B.S. degree from the Department of Computer Science and Technology, Shanghai Jiao Tong University, in 2015. She is currently a Researcher with Tencent AI Lab. Her research interests include 3D face reconstruction and image synthesis.
\end{IEEEbiography}
\vspace{-4em}

\begin{IEEEbiography}[{\includegraphics[width=0.8in,height=1.25in,clip,keepaspectratio]{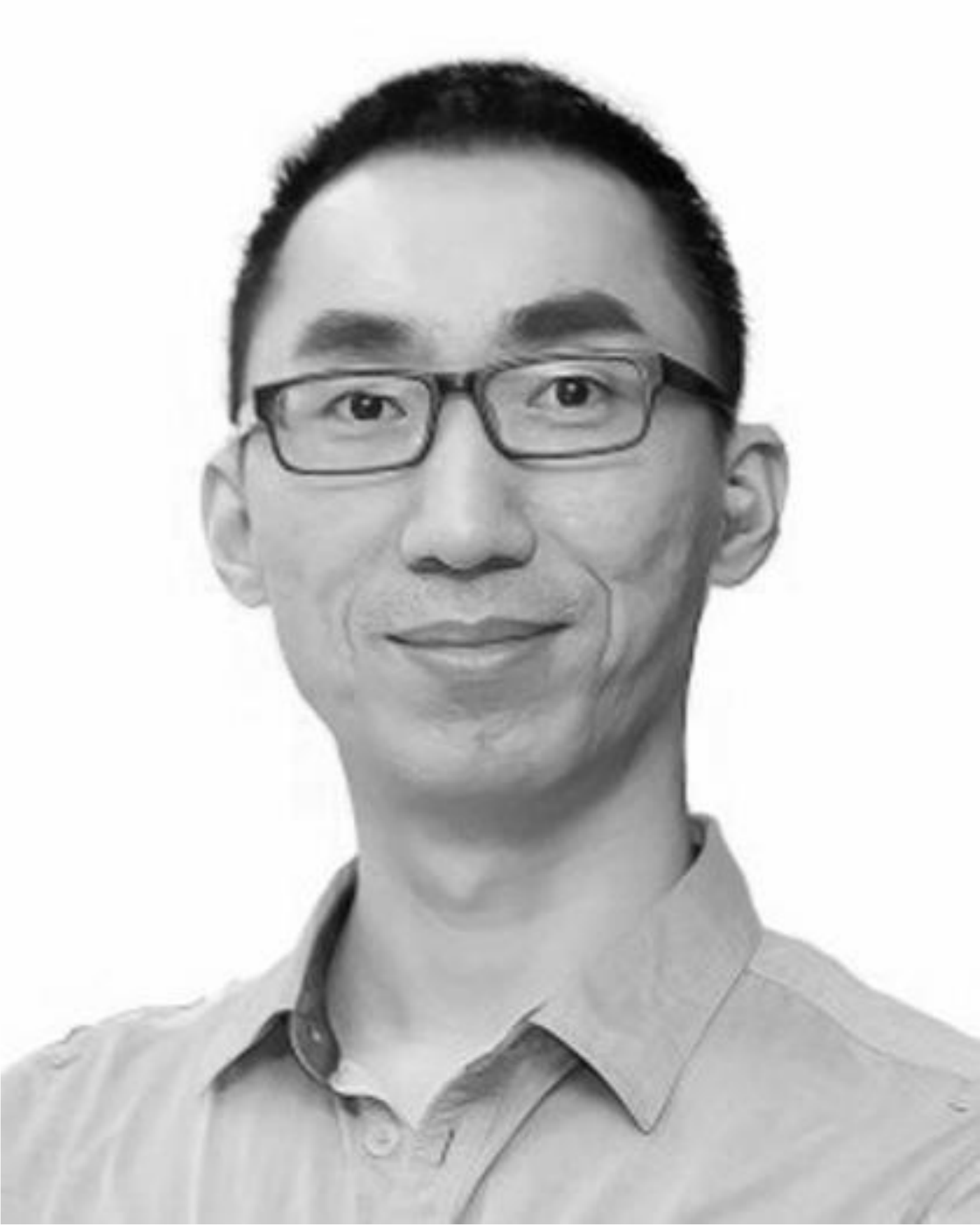}}]{Linchao Bao} received the M.S. degree in pattern recognition and intelligent systems from the Huazhong University of Science and Technology, Wuhan, China, and the Ph.D. degree in computer science from the City University of Hong Kong in 2015. From November 2013 to August 2014, he was a Research Intern with Adobe Research. From January 2015 to June 2016, he worked as an Algorithm Engineer with DJI. He is currently a Principal Research Scientist with Tencent AI Lab. His research interests include computer vision and graphics.
\end{IEEEbiography}
\vspace{-3em}

\begin{IEEEbiography}[{\includegraphics[width=0.8in,height=1.25in,clip,keepaspectratio]{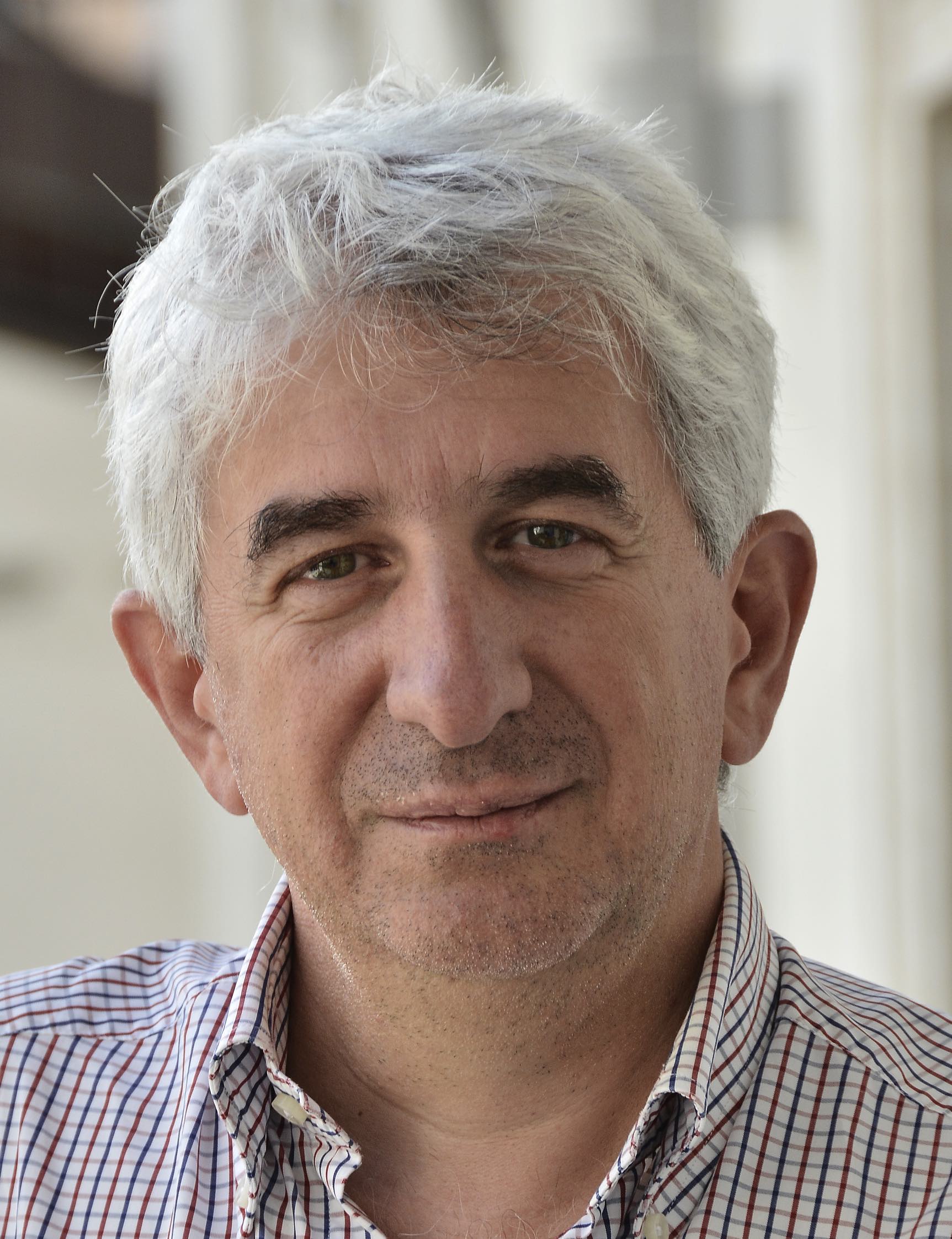}}]{Nicu Sebe} (Senior Member, IEEE) is a Professor with the University of Trento, Italy, leading the research in the areas of multimedia information retrieval and human behavior understanding. Dr. Sebe was the General Co-Chair of the IEEE FG 2008 and the Association for Computing Machinery
(ACM) Multimedia 2013. He was a Program Chair of the ACM Multimedia 2011 and
2007, European Conference on Computer Vision
(ECCV) 2016, International Conference on Computer
Vision (ICCV) 2017, and International Conference
on Pattern Recognition (ICPR) 2020. He is a General Chair of ACM
Multimedia 2022 and a Program Chair of ECCV 2024. He is a fellow of the
International Association of Pattern Recognition (IAPR).
\end{IEEEbiography}
\vspace{-3em}

\begin{IEEEbiography}[{\includegraphics[width=0.8in,height=1.25in,clip,keepaspectratio]{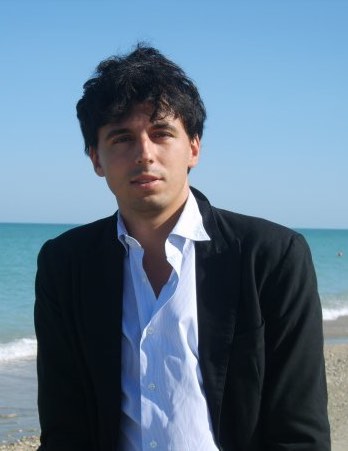}}]{Bruno Lepri} received the Ph.D. degree in computer science from the University of Trento, Italy, in 2009. He currently leads the Mobile and Social Computing Laboratory (MobS) at Fondazione Bruno Kessler (FBK), Trento, Italy. From 2010 to 2013, he was a Joint Postdoctoral Fellow with the MIT Media Laboratory, Boston, MA, USA, and FBK. His research interests include computational social science, human behavior understanding, and machine learning. He received the James Chen Annual Award for the Best UMUAI Paper, and the Best Paper Award from ACM Ubicomp in 2014. He served as Area Chair of ACM Multimedia 2020 and 2021.
\end{IEEEbiography}
\vspace{-3em}

\begin{IEEEbiography}[{\includegraphics[width=0.8in,height=1.25in,clip,keepaspectratio]{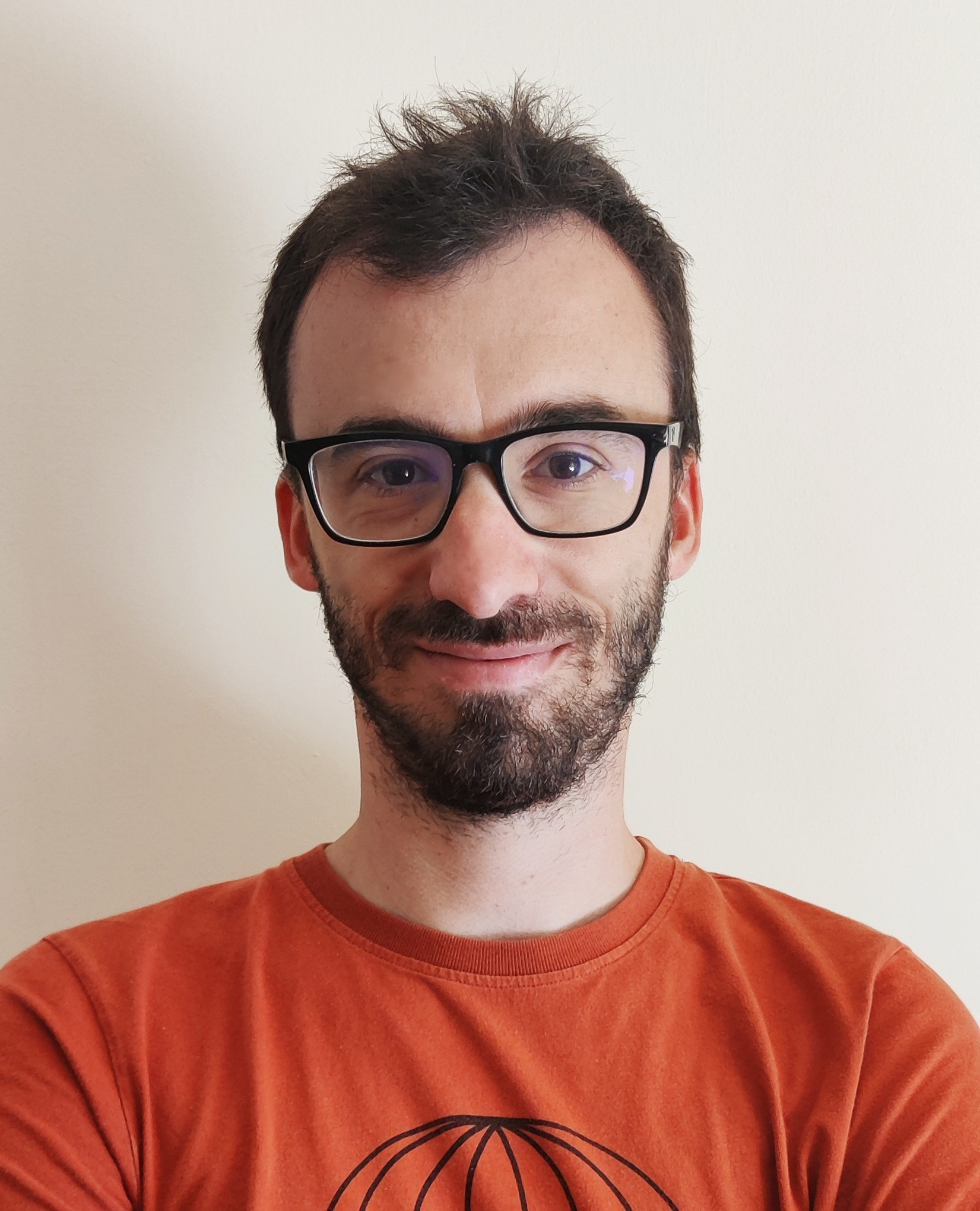}}]{Marco De Nadai} is a Research Scientist at Fondazione Bruno Kessler, Trento, Italy. His expertise lies in Human behavioural understanding through the analysis of multi-modal data, computer vision, and unsupervised learning. Marco holds a PhD and a Master degree in Computer Science at the University of Trento, Italy, where he was awarded for the best Master and best PhD student. He collaborated with numerous international institutions including MIT, MIT Media Lab, Nokia Bell Labs, Vodafone Group and Data-pop Alliance.
\end{IEEEbiography}

\vfill

\appendix
\renewcommand\thefigure{\arabic{figure}}
\setcounter{figure}{10}

\section{Additional details}

\subsection{Evaluation metrics} 

\begin{itemize}[leftmargin=*,noitemsep,topsep=0pt]	
    \item \emph{FID.} We adopt the Fr\'echet Inception Distance (FID)~\cite{martin2017fid} to compare generated images with real images. To calculate FID score, we randomly sample 20K images from FFHQ dataset as real images. A lower FID score indicates a lower discrepancy between the image quality of the real and generated images. We collect the comparisons on four semantic editing tasks, namely Age (A), Gender (G), Expression (EP) and Eyeglasses (EG), and report the average FID values among all tasks. 
    \item \emph{LPIPS.} We evaluate the diversity among generated images with the Learned Perceptual Image Patch Similarity (LPIPS)~\cite{zhang2018unreasonable}. The LPIPS distance is defined as the $L_2$ distance between the features extracted by a deep learning model of two images. Following~\cite{ huang2018multimodal, zhu2017toward}, we randomly select 1K input images and translate them to different domains. We generate 10 images for each input image for each domain translation and evaluate the average LPIPS distance between the 10 generated images. Finally, we get the average of all distances. Higher LPIPS distance indicates better diversity among the generated images. 
    \item \emph{Accuracy.} We use the pre-trained attribute classifier on CelebA to evaluate the accuracy of semantics in the translated images. A higher accuracy refers to higher correctness of expected attributes appearing in the synthesized images during the translation. The averaged accuracy of different semantic attributes is named as \emph{mAcc}.
\end{itemize}

\section{Additional results}

\subsection{Results per type of translation}

To better assess the results of ISF-GAN we here provide our results divided per type of translation. 
\Cref{tab:result-ACC} shows that we achieve the best results in all the tasks but in the expression task in Set2. However, ISF-GAN improves results by a large margin in all the other tasks.

\Cref{tab:result-FID} shows instead the results on FID, where once again we achieve the best results.

\begin{table}[!ht]
    \centering
    \small
    \begin{tabularx}{\columnwidth}{@{}X c rrrr@{}}
    \toprule
    \textbf{Method} & \textbf{Dataset} &\textbf{A} &  \textbf{G} & \textbf{EP} & \textbf{EG} \\
    \midrule
        InterFaceGAN~\cite{shen2020interfacegan} & \multirow{2}{*}{\textit{Set}$_1$} & 85.9 & \textbf{96.5} & 94.6 & \textbf{99.7} \\
        Ours &  & \textbf{91.8} & 95.0 & \textbf{96.3} & 98.7 \\
        \midrule
        StyleFlow~\cite{abdal2020styleflow} & \multirow{3}{*}{\textit{Set}$_2$} & 82.4 & 70.3 & 74.5 & 61.7 \\
        InterFaceGAN~\cite{shen2020interfacegan} & & 74.8 & 71.6 & \textbf{99.3} & 70.1 \\
        Ours &  & \textbf{96.1} & \textbf{93.7} & 96.6 & \textbf{99.7} \\ %
        \bottomrule
    \end{tabularx}
    \caption{Classification accuracy (\%) in latent space with respect to different attributes. A, G, EP and EG refer to \textit{age}, \textit{gender}, \textit{expression} and \textit{eyeglasses} translations, respectively. %
    }
    \label{tab:result-ACC}
\end{table}

\begin{table}[!ht]
    \centering
    \small
    \begin{tabularx}{\columnwidth}{@{}X c rrrr rrrr@{}}
    \toprule
    \textbf{Method} & \textbf{Dataset} & \textbf{A} & \textbf{G} & \textbf{EP} & \textbf{EG} \\
    \midrule
        InterFaceGAN~\cite{shen2020interfacegan} & \multirow{2}{*}{\textit{Set}$_1$} & 58.55 & 48.75 & 41.18 &  55.10 \\
        Ours & & \textbf{31.48} & \textbf{28.86} & \textbf{27.48} & \textbf{29.96} \\
        \midrule
        StyleFlow~\cite{abdal2020styleflow} & \multirow{3}{*}{\textit{Set}$_2$} & 56.65 & 65.81 & 51.29 & 50.36 \\ 
        InterFaceGAN~\cite{shen2020interfacegan} &  & 30.48 & 30.67 & 29.23 & 29.02\\
        Ours & &  \textbf{27.98} & \textbf{21.43} & \textbf{22.26} & \textbf{24.07} \\
        \bottomrule
    \end{tabularx}
    \caption{Image quality on FID in latent space with respect to different attributes. A, G, EP and EG refer to \textit{age}, \textit{gender}, \textit{expression} and \textit{eyeglasses} translations, respectively. %
    }
    \label{tab:result-FID}
\end{table}

\subsection{Qualitative results}

\begin{figure}[ht]	
	\renewcommand{\tabcolsep}{1pt}
	\centering
	\begin{tabular}{c}
	    \includegraphics[width=\linewidth]{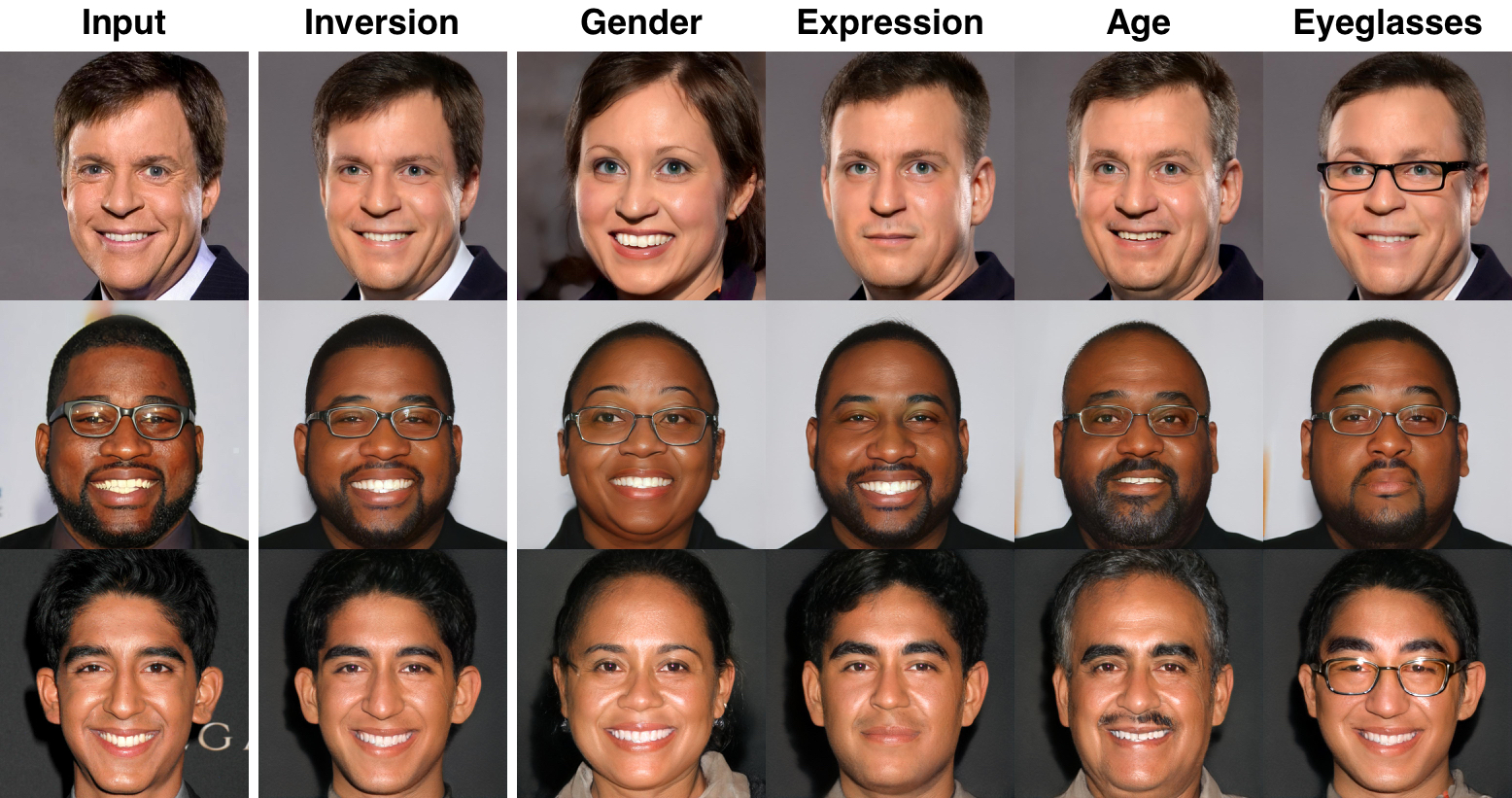}
	\end{tabular}
	\caption{Manipulations of real images from CelebA-HQ~\cite{karras2018progressive} through ISF-GAN  and an image embedding method~\cite{abdal2019image2stylegan}.
	}
	\label{Fig:real-image-appendix}
	\vspace{-0.5em}
\end{figure}

\Cref{Fig:real-image-appendix} shows more results on manipulating real images from the CelebA-HQ dataset. 

\begin{figure}[!ht]	
	\renewcommand{\tabcolsep}{1pt}
	\centering
	\begin{tabular}{c}
	    \includegraphics[width=0.96\linewidth]{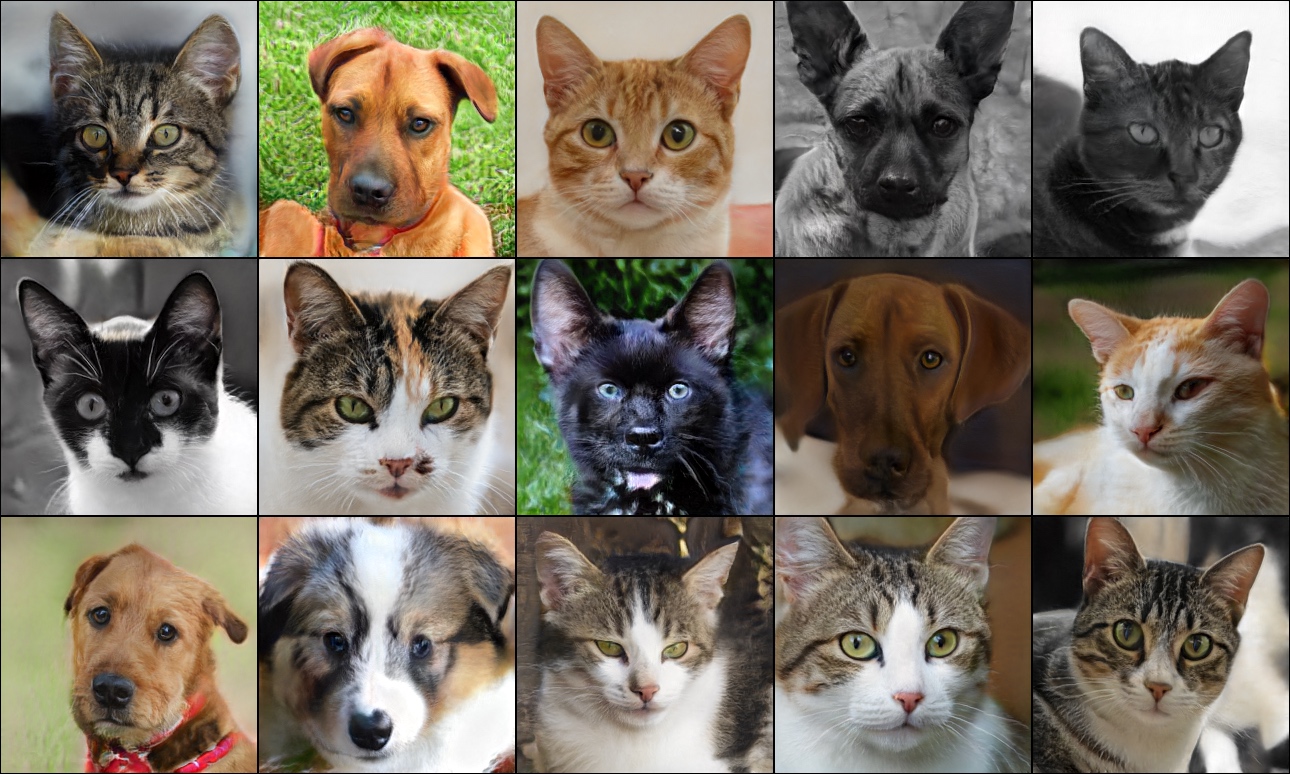}
	\end{tabular}
	\caption{Random sampling in the latent space of pretrained StyleGAN v2 on Cat$\leftrightarrow$Dog dataset~\cite{choi2019stargan}.
	}
	\label{Fig:cat2dog}
\end{figure}

\Cref{Fig:cat2dog} shows some sampled images from StyleGAN v2 trained on Cat$\leftrightarrow$Dog dataset~\cite{choi2019stargan}.

\begin{figure*}[!ht]	
	\renewcommand{\tabcolsep}{1pt}
	\centering
	\begin{tabular}{c}
	    \includegraphics[width=\linewidth]{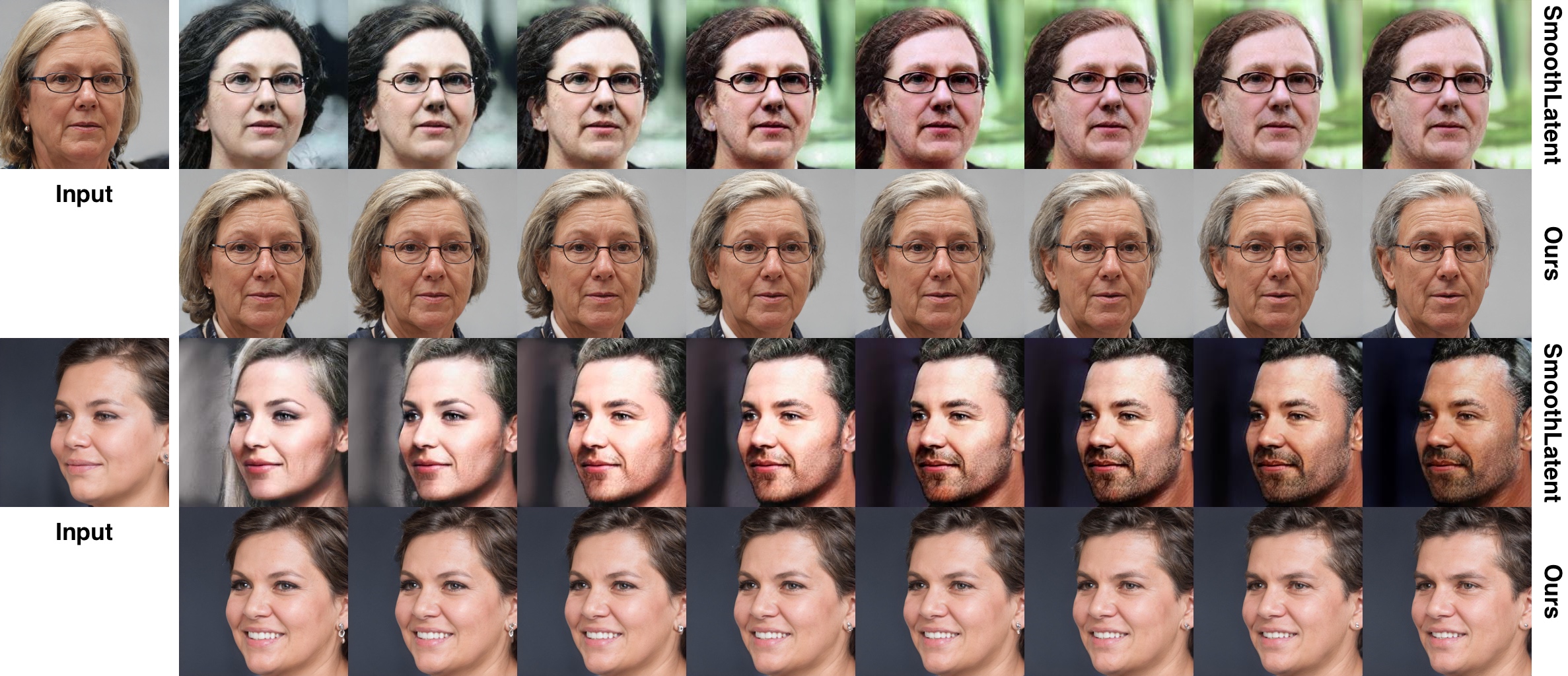}
	\end{tabular}
	\caption{Visual comparisons between the proposed ISF-GAN and SmoothLatent~\cite{Liu_2021_CVPR} on gender interpolation. Although SmoothLatent can synthesize smooth interpolations, ISF-GAN generates more realistic images.
	}
	\label{Fig:vs-smoothlatent}
\end{figure*}

\begin{figure*}[!ht]	
	\renewcommand{\tabcolsep}{1pt}
	\centering
	\begin{tabular}{c}
	    \includegraphics[width=0.98\linewidth]{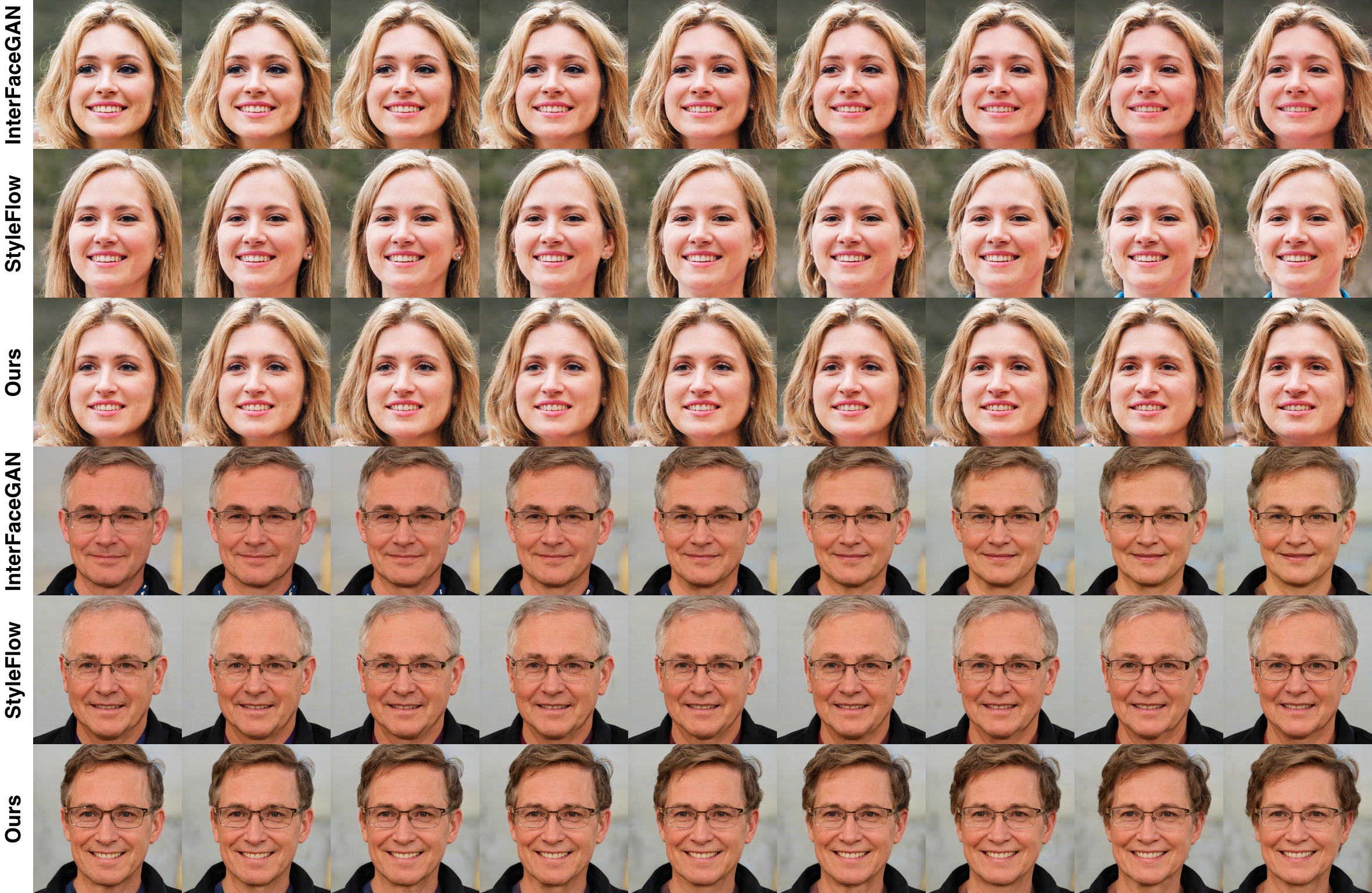}
	\end{tabular}
	\caption{Comparisons on gender interpolations between InterFaceGAN~\cite{shen2020interpreting}, StyleFlow~\cite{abdal2020styleflow} and our proposed method tested on \textit{Set}$_2$. Compared to InterFaceGAN and StyleFlow, our method preserves better the face identity along the interpolations.} 
	\label{Fig:interpolation-set2}
\end{figure*}

\begin{figure*}[ht]	
	\renewcommand{\tabcolsep}{1pt}
	\centering
	\begin{tabular}{c}
	    \includegraphics[width=0.98\linewidth]{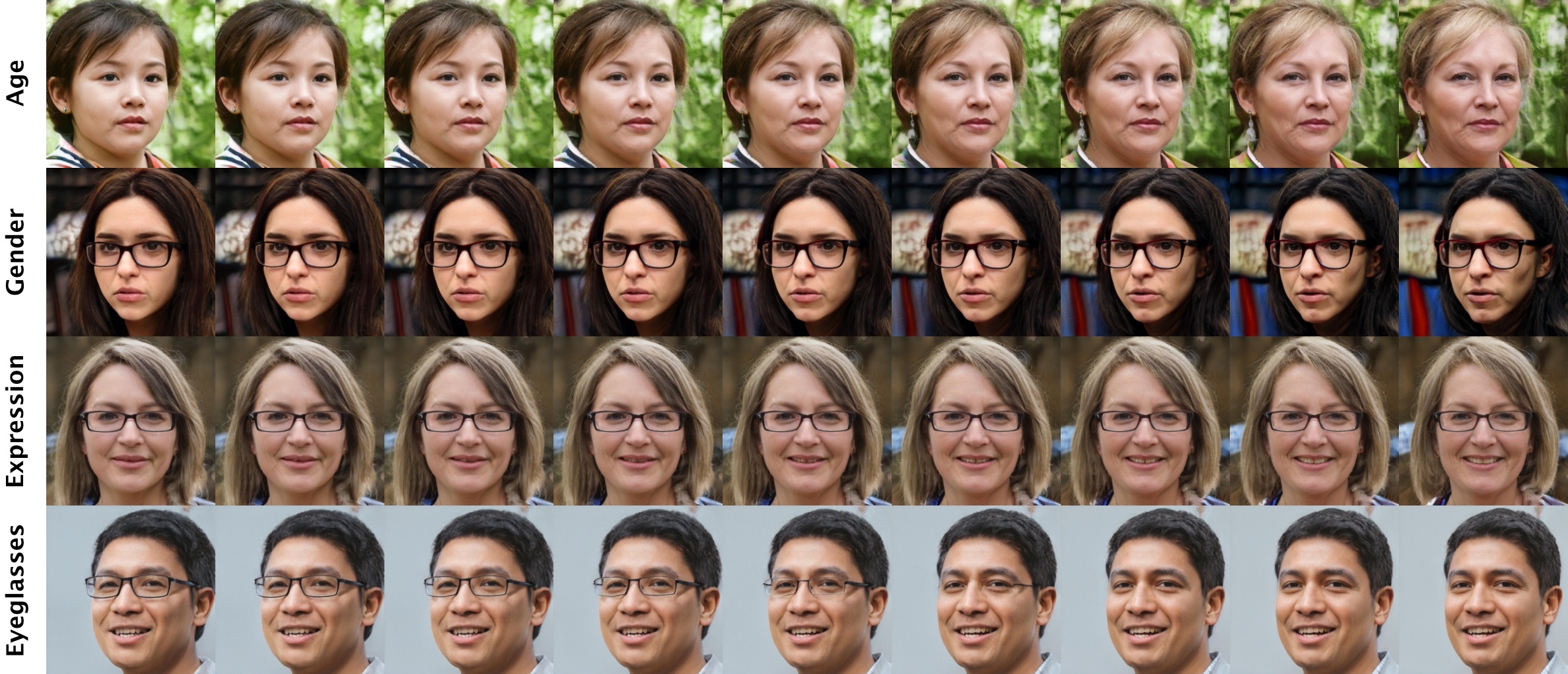}
	\end{tabular}
	\caption{Smooth inter-domain interpolations of our proposed ISF-GAN on various face attributes.
	}
	\label{Fig:smooth-interpolation}
\end{figure*}

\Cref{Fig:vs-smoothlatent}, \Cref{Fig:interpolation-set2} and \Cref{Fig:smooth-interpolation} show the smoothness of our results.

\end{document}